\documentclass[letterpaper, 10 pt, conference]{ieeeconf}

\IEEEoverridecommandlockouts
\usepackage{cite}
\usepackage{booktabs}
\usepackage{pifont}
\newcommand{\cmark}{\ding{51}}
\newcommand{\xmark}{\ding{55}}

\let\labelindent\relax
\usepackage{enumitem}
\usepackage{bm}
\usepackage{amsfonts,amssymb,amsmath}
\usepackage{graphicx}
\usepackage{capt-of}
\usepackage{cuted}
\usepackage{float}
\usepackage{multirow}
\usepackage[percent]{overpic}
\usepackage{tikz}
\usepackage{siunitx}
\usepackage{makecell}
\usepackage[table]{xcolor}
\usepackage[caption=false,font=normalsize,labelfont=sf,textfont=sf]{subfig}
\usepackage[normalem]{ulem}
\usepackage{url}

\usepackage{textcomp}

\makeatletter
\let\NAT@parse\undefined
\makeatother

\usepackage[colorlinks=true,linkcolor=blue,citecolor=blue,urlcolor=blue]{hyperref}

\newcommand{\gray}[1]{\textcolor{gray!60}{#1}}

\begin{document}
\bstctlcite{IEEEexample:BSTcontrol}

\title{
Learning Asynchronous Upper-body Task-space Trajectory Tracking Policy for Humanoid Robots
}

\author{
Yumeng Liu$^{1}$, Dongqi Wang$^{1}$, Jiyu Yu$^{1}$, Yijun Fan$^{1}$, Rong Xiong$^{1}$, Yue Wang$^{1}$\\
$^{1}$Zhejiang University\\
}

\maketitle

\vspace*{-1cm}

\begin{abstract}
High-level humanoid planners often output structurally sparse task-space trajectories at low update rates, whereas whole-body controllers run at high frequency. This creates \textit{temporal asynchrony} between the planning and execution, and \textit{structural incompleteness} for full-body control. We propose an asynchronous upper-body task-space tracking framework for humanoids. A student policy is initialized by teacher–student distillation, conditioned on the full cached future trajectory and an execution-time index, and trained with a sliding-window global reward to learn to handle planning-frame mismatch under asynchronous updates without using the estimated drift for online reference correction. For task-specific post-training, an MPC module completes sparse references into floating-base and upper-body guidance, while action- and FK-level self-guidance constrain policy drift. Simulation and Unitree G1 hardware experiments show improved tracking under low update rates, stronger performance than synchronous and decoupled baselines, and safer adaptation to out-of-distribution motions.
\end{abstract}

\section{Introduction}
\label{sec:intro}

\begin{figure}[t]
\vspace{-0.6cm}
\centering
\includegraphics[width=\linewidth,trim=0 0 5pt 0,clip]{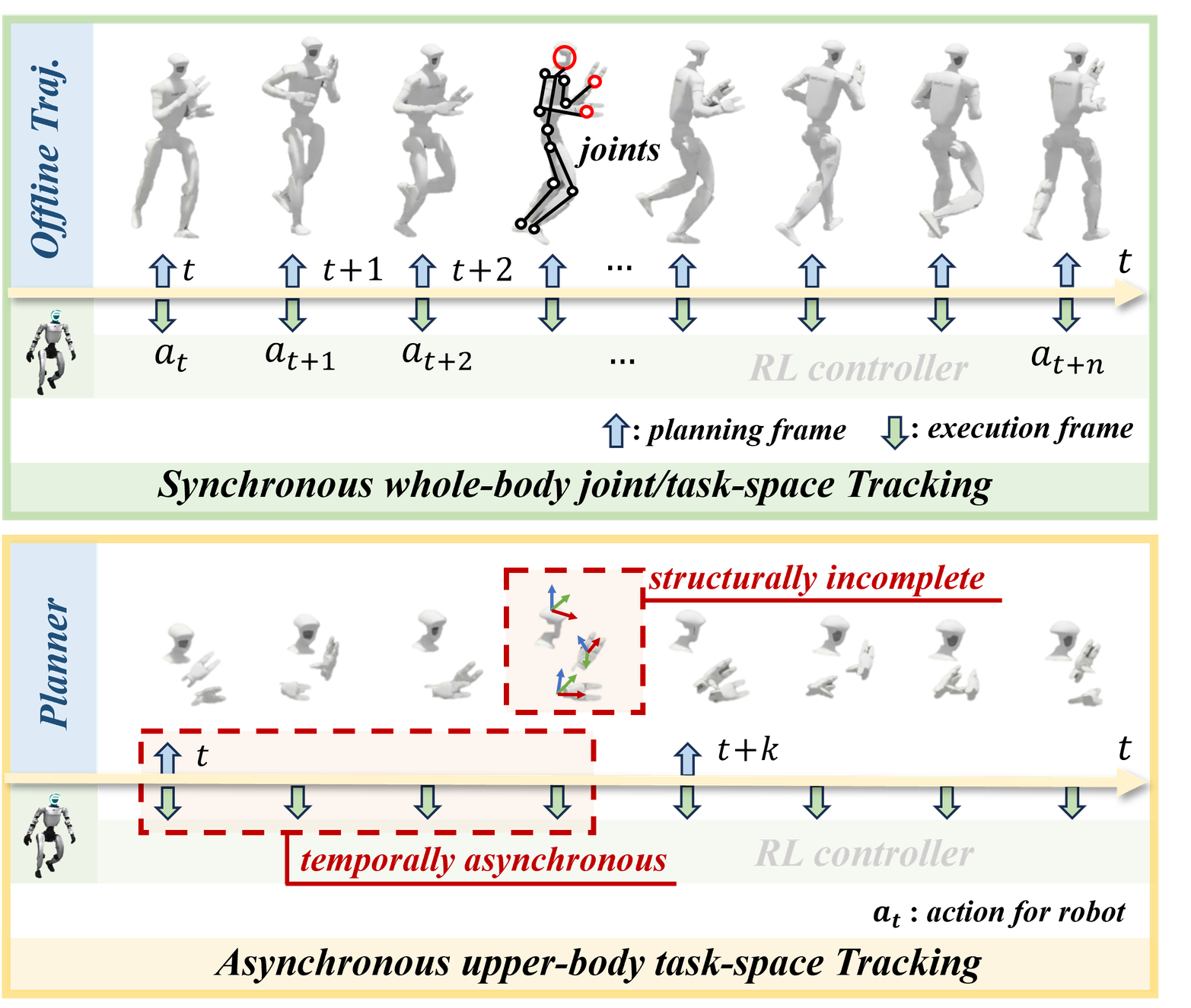}
\vspace{-0.5cm}
\caption{Execution-interface gap between task-level planners and low-level humanoid controllers.
Top: synchronous tracking uses dense, frame-aligned whole-body references.
Bottom: low-rate planners provide structurally sparse trajectories, while high-rate controllers execute cached commands.
This creates two challenges:
(1) \emph{temporal asynchrony}, causing reference-frame mismatch between planning and execution;
and (2) \emph{structural incompleteness}, leaving an under-constrained problem for a 29-DoF humanoid.}
\vspace{-0.4cm}
\label{fig:two_problem}
\end{figure}

Humanoid control has shifted from servo‑level tracking to task‑level execution. Instead of fixed trajectories or human teleoperation, commands are now generated by high‑level planners. This makes the control policy problem an asynchronous upper‑body task‑space tracking problem. The problem introduces two challenges as shown in Fig.~\ref{fig:two_problem}. First, the planner runs at a low frequency (1–10Hz), while the low‑level controller must operate at a much higher frequency ($\sim$50Hz) to maintain stability. Each planner output is executed over multiple control steps. We refer to this planner--controller rate mismatch as \textit{temporal asynchrony}. Second, the reference includes only the end-effector trajectory, the hands and head, in this work. We refer to this structural incompleteness as \textit{structural sparsity}, which leaves the whole-body control problem under-constrained.


Teacher–student distillation is common for control policy learning against structural incompleteness. A teacher policy is trained with a privileged whole-body dataset; a student policy is then distilled from the teacher using only the sparse upper‑body reference~\cite{h2o,gmt,hover}. This paradigm works well when references are updated synchronously. Under asynchronous updates, however, the policy experiences a frame mismatch: the reference remains tied to its frame at planning time, while the robot’s frame changes over multiple control steps, as mentioned above. Since the frame drift is not directly observable, existing methods usually rely on two types of remedies. The first explicitly estimates the missing frame transform~\cite{hiwet,hero}, which breaks end‑to‑end learning and introduces additional estimation errors. The second converts the sparse reference into local incremental velocity commands~\cite{grootn1, agibot}. Although this reduces the timing mismatch, the incremental errors can accumulate over time and lead to severe drift.


When transferring a pre‑trained policy to a specific task, zero‑shot execution is rarely sufficient. Fine‑tuning on task‑specific data is typically required\cite{lift}. This is easy for whole‑body tracking, but becomes challenging when only an upper‑body trajectory is available. The structurally incomplete reference complicates task‑specific post‑training: for whole‑body tracking, the same distillation can be applied in both pre‑training and post‑training. However, when a new task provides only an upper‑body trajectory, the privileged teacher, which expects whole‑body states, is neither usable nor fine‑tunable. The student policy loses its imitation supervision and must rely solely on reinforcement learning (RL) with sparse task rewards. This under‑constrains the high‑dimensional humanoid policy, leading to unstable training and suboptimal performance.


\begin{figure*}[t]
    \centering
    \includegraphics[width=\hsize]{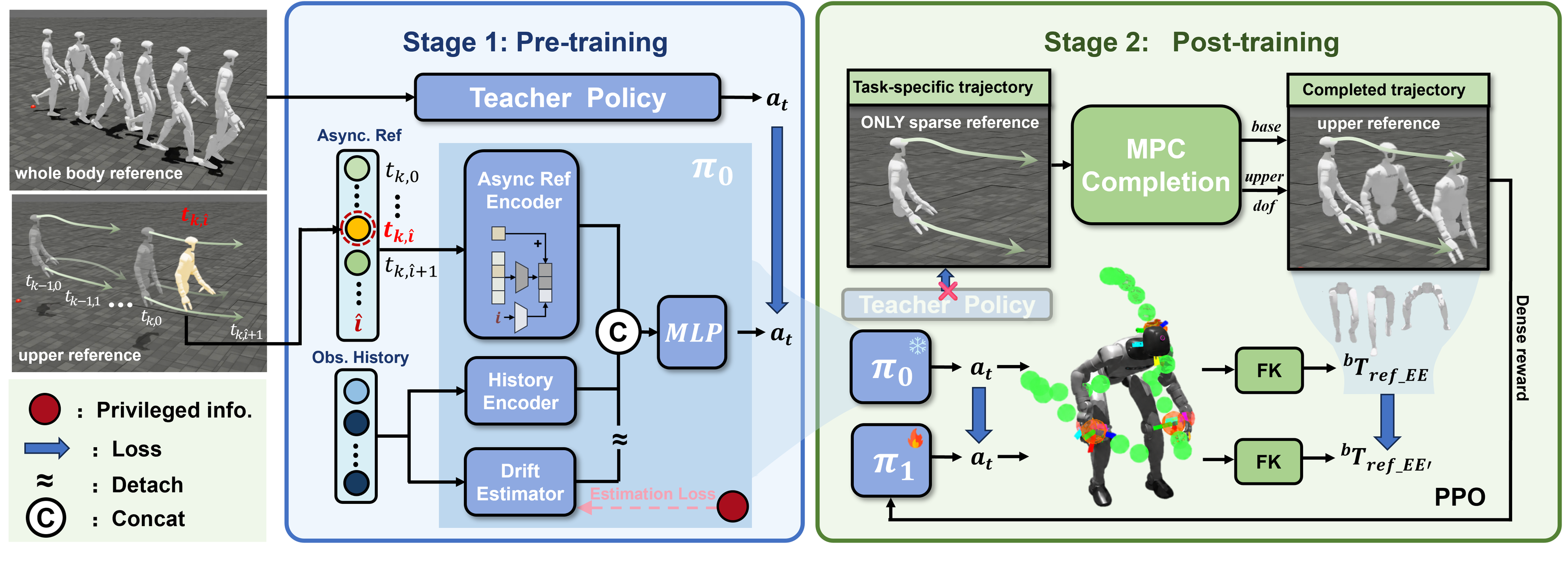}
    \vspace{-0.8cm}
    \caption{
    Overview of the proposed asynchronous sparse trajectory tracking and post-training framework.
    \textbf{Stage 1: Pre-training.} Pretrain the student policy ($\pi_0$) via teacher--student distillation with entire sparse references, a time index, history encoding, and auxiliary drift estimation.
      \textbf{Stage 2: Post-training.} 
      Adapts $\pi_0$ to task-specific sparse trajectories using MPC-completed base and upper-body references. The pretrained policy $\pi_0$ is frozen and serves as a pseudo-teacher, providing action-level
      and FK-level self-guidance during RL post-training for $\pi_1$.}
    \vspace{-0.3cm}
    \label{fig:overview}
\end{figure*}

To address both challenges, we propose a learning framework for asynchronous upper‑body trajectory tracking. For temporal asynchrony, we introduce a time‑indexed conditioning policy that implicitly aligns the robot’s motion with the reference without explicitly estimating the drift base-frame transform, enabling end-to-end learning. In addition, a sliding‑window global reward further penalizes error accumulation over each planning interval. For structural incompleteness, we complete the sparse upper‑body trajectory using MPC and incorporate the completed trajectory into both the policy observation and the RL reward. This stabilizes RL post‑training. A self‑guidance regularization additionally constrains the lower‑body behavior, preventing distributional drift. Experiments in the simulation and on a real humanoid robot show that the proposed framework outperforms the baselines and comparative methods. Our contributions are:
\begin{itemize}
    \item A learning framework that bridges low-frequency, upper-body-only planner outputs to whole-body humanoid control, handling asynchronous and incomplete references.
    \item A time index conditioned policy with a sliding‑window global reward for implicit spatiotemporal alignment.
    \item A post‑training method that integrates MPC‑based self‑completion with self‑guidance regularization.
    \item Simulation and real‑world validation on a humanoid robot, demonstrating stable adaptation to novel motions.
\end{itemize}


\section{Related Works}

\subsection{Task-level Planner Interfaces}

Recent humanoid systems are shifting from servo-level tracking to task-level action generation. VLA models and humanoid foundation policies can generate actions from language, vision, proprioception, and demonstrations~\cite{grootn1,egohumanoid,psi0,sonic}. Different planner interfaces provide different levels of embodiment generality and execution guidance. Controller-native latent or action-token interfaces provide rich motion priors but tightly couple the upstream planner to a specific controller and embodiment. For example, GR00T with SONIC predicts latent motion tokens decoded directly by the pretrained whole-body controller~\cite{grootn1,sonic}. Explicit whole-body interfaces reduce execution ambiguity but remain morphology-specific: VLK predicts root and whole-body joint trajectories, while $\Psi_0$ predicts upper-body joint actions and structured lower-body commands executed by AMO~\cite{VLK,psi0}. In contrast, sparse task-space interfaces support heterogeneous embodiments and scalable data collection. TrajBooster uses 6D end-effector trajectories as a morphology-agnostic interface, while HuMI obtains task-space keypoint trajectories from robot-free human demonstrations~\cite{TrajBooster,humi}. Such portability leaves whole-body realization more under-specified; our work addresses this setting under low-rate asynchronous planning.

\subsection{Asynchronous Trajectory Tracking}
\vspace{-0.3em}
Existing work usually handles the planner--controller gap by making low-rate task commands executable for a high-rate controller. A common approach is to buffer a future action chunk and execute the waypoint that matches the current timestep~\cite{grootn1,homie,humi}. This resolves the rate mismatch between planning and control. In this case, the selected waypoint is generated in the planning-time frame, but is executed as an instantaneous target under the current floating-base state, leaving the geometric mismatch unresolved. 
Another line of work use local incremental commands, such as base velocities or short-horizon relative keypoint motions~\cite{egohumanoid,psi0,humi,agibot}. They compress a future trajectory into step-wise local updates, so the long-horizon positional structure is weakened and integration errors may accumulate over time. 
More geometric treatments reconstruct or refresh the tracking target during execution. For example, HiWET formulates end-effector tracking in a world frame to maintain spatial consistency~\cite{hiwet}, while HERO improves tracking with learned FK/odometry, goal adjustment, and periodic replanning~\cite{hero}. These methods reduce geometric drift, but they usually rely on global localization, state estimation, or online correction. Also, their odometry always assumes planted feet, which fails under the moving-base trajectories considered in this work. The estimator and controller are also often optimized separately. 
In contrast, our policy observes the entire future trajectory segment together with an execution‑phase index. This allows the controller to learn the consistency between the planning‑time reference and the current state in an end-to-end manner without explicit frame estimation.

\subsection{Tracking Policy Post-training}
Recent humanoid control works show that pretrained whole-body controllers often require post-training when the task or reference distribution changes~\cite{refinedp,exbody2,fromw1}. 
LIFT studies efficient fine-tuning of pretrained humanoid policies in new environments and out-of-distribution(OOD) tasks~\cite{lift}. AHC further follows a distillation-then-adaptation pipeline, where a multi-behavior controller is first obtained by behavior distillation and then improved through reinforced fine-tuning~\cite{ahc}. 
However, these methods do not directly address controller post-training from low-rate upper-body-only structurally sparse trajectories, where the floating-base and lower-body motions are unspecified. In teacher--student trackers, the dense teacher supervision is unavailable when adapting to new sparse references and direct RL fine-tuning with sparse rewards is also prone to policy drift or forgetting~\cite{fast,ppf}. We therefore use self-guided post-training, where the pretrained student acts as a pseudo-teacher and MPC-completed references provide auxiliary rewards.


\section{Overview}
\label{sec:overview}

We propose a two‑stage framework to address asynchronous upper‑body task‑space tracking, as illustrated in Fig.~\ref{fig:overview}. In Stage~1 (Sec.~\ref{sec:problem}), the pre‑training stage, a whole‑body motion dataset is available. We adopt teacher–student distillation and introduce ASYNC‑3PT to handle temporal asynchrony. In Stage~2 (Sec.~\ref{sec:ASYNC-CA}), the post‑training stage, only task‑specific upper‑body trajectories are provided. We therefore extend ASYNC‑3PT to ASYNC‑CA by additionally considering the base motion beyond the head and hands. To this end, we employ MPC to complete the base trajectory given the upper‑body reference, which augments the policy observation and enables self‑adaptation under structural incompleteness. Together, the two stages form a complete framework for the asynchronous tracking problem.
\section{\textbf{ASYNC-3PT}: Time-Indexed Conditioning for Asynchronous Tracking}
\label{sec:problem}

\begin{figure}
    \centering
    
    \includegraphics[width=\linewidth]{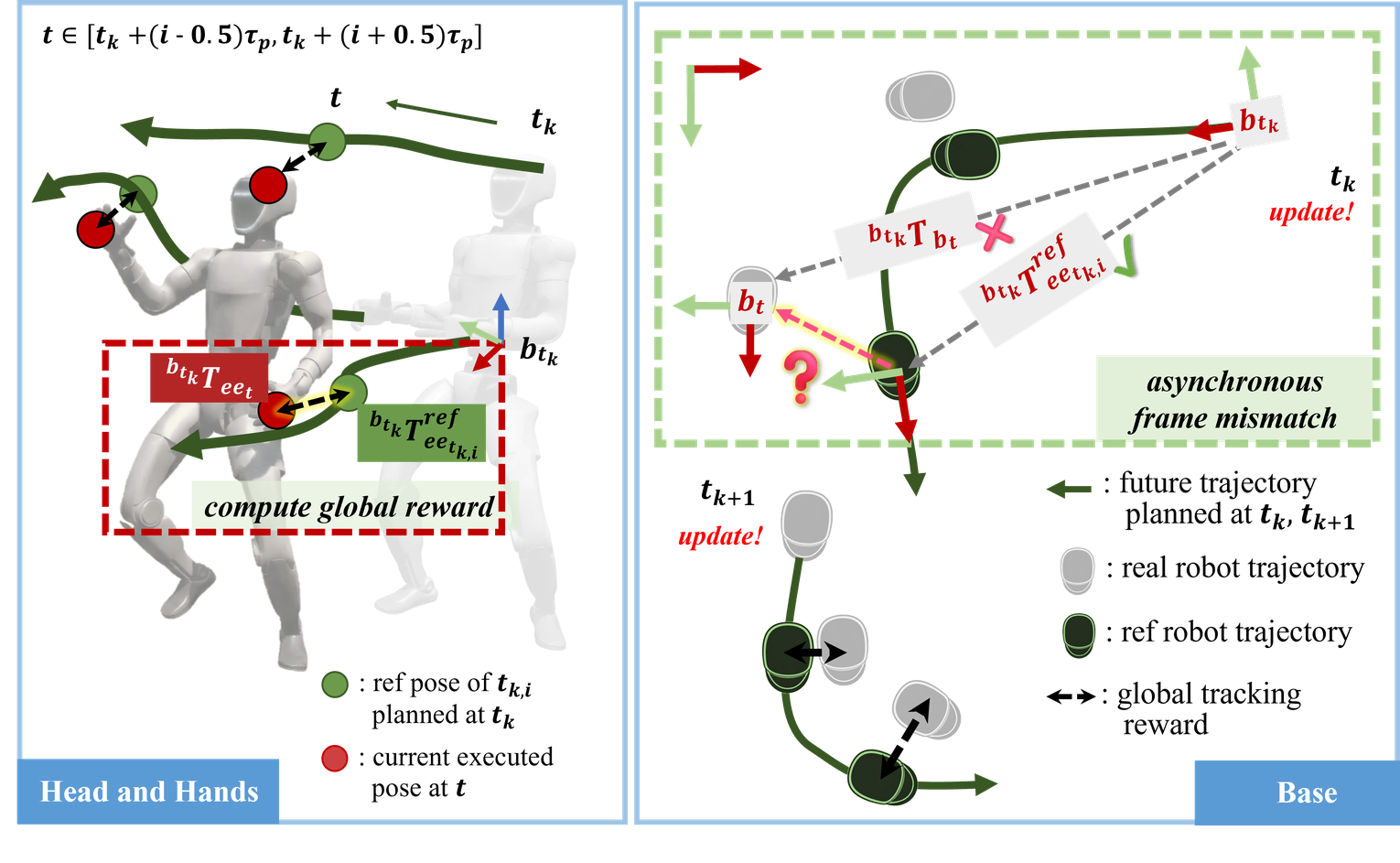}
    \vspace{-0.8cm}
    \caption{\textbf{Asynchronous frame mismatch and sliding-window global reward.}
\textbf{Base (right):} the base drifts from $b_{t_k}$ to $b_t$ between updates,
so the transform ${}^{b_{t_k}}\mathbf{T}_{b_t}$ is unavailable, causing the
mismatch of the reference frame. \textbf{Head and Hands (left):} we keep the
cached reference and executed pose in the same frame $b_{t_k}$, and penalize
their global discrepancy over a sliding window.}
    \label{fig:global_rwd}
    \vspace{-0.2cm}
\end{figure}

This section formalizes Stage~1 of our framework.
We first define and derive the frame mismatch caused by asynchronous execution, and then formulate the ASYNC-3PT policy to consider the asynchronous time.

We assume that, a floating-base humanoid is commanded by a high-level planner whose update rate is lower
than the control frequency. At each planning time $t_k$, the planner produces a finite-horizon
reference for task-relevant end-effectors, such as the head and dual hands (Fig.~\ref{fig:global_rwd},left) . Let $t_{k,i}=t_k+i\tau$ denote the timestamp of the $i$-th future waypoint, where $i=0,\ldots,N_{\mathrm{ref}}-1$ and $\tau$ is the waypoint spacing. The planner provides the relative end-effector motion ${}^{ee_{t_k}}\mathbf{T}^{ref}_{ee_{t_{k,i}}} \in SE(3)$, which describes the future pose change of the end-effector with respect to its pose at the planning
time $t_k$.
Using the end-effector pose from forward kinematics,
\begin{equation}
{}^{b_{t_k}}\mathbf{T}_{ee_{t_k}} = \mathrm{FK}(\mathbf{q}_{t_k}),
\end{equation}
where $\mathbf{q}_{t_k}$ denotes the joint positions, we express the sparse future trajectory in the
planning-time base frame as
\begin{equation}
{}^{b_{t_k}}\mathbf{T}^{ref}_{ee_{t_{k,i}}}
=
{}^{b_{t_k}}\mathbf{T}_{ee_{t_k}}
{}^{ee_{t_k}}\mathbf{T}^{ref}_{ee_{t_{k,i}}}.
\end{equation}
The sparse reference is therefore
\begin{equation}
\label{reftraj}
\mathcal{T}_k =
\left\{
{}^{b_{t_k}}\mathbf{T}^{ref}_{ee_{t_{k,i}}}
\right\}_{i=0}^{N_{\mathrm{ref}}-1}.
\end{equation}
It is a low-rate, structurally sparse reference defined in the planning-time base frame $b_{t_k}$, covering
only a subset of end-effectors.

Unless otherwise specified, each cached reference $\mathcal{T}_k$ contains
$N_{\mathrm{ref}}=20$ future waypoints with spacing $\tau=0.1$~s.
The controller operates at $\Delta t_c=0.02$~s (50~Hz). We use
$\Delta t_k=1$~s during training.

\subsection{Asynchronous Execution and Frame Mismatch}
\label{sec:async_model}

The low-level controller runs at a higher frequency than the planner. 
As a result, each trajectory $\mathcal{T}_k$ is tracked over multiple control steps until the next
reference update at $t_{k+1}$. During this interval, the floating base keeps moving, while the
reference trajectory remains defined in the planning-time base frame $b_{t_k}$. At execution time $t>t_k$, the robot observations are expressed in the current base frame $b_t$. 
Ideally, one should transform the planned reference from $b_{t_k}$ to $b_t$:
\begin{equation}
{}^{b_t}\mathbf{T}^{ref}_{ee_{t_{k,i}}}
=
{}^{b_t}\mathbf{T}_{b_{t_k}}
{}^{b_{t_k}}\mathbf{T}^{ref}_{ee_{t_{k,i}}}.
\end{equation}

However, the relative transform ${}^{b_t}\mathbf{T}_{b_{t_k}}$ is not directly available without
reliable global localization, leading to mismatch of the reference frame.
As a result, the planned reference and the current robot state are represented in different frames, so the task-space tracking error cannot be directly measured, making asynchronous tracking challenging.

\subsection{Asynchronous Time-indexed Conditioning}
\label{sec:async}

To handle this mismatch in an end-to-end manner, we introduce a sparse end-effector asynchronous
controller, denoted as ASYNC-3PT.
The key idea is to condition the policy on the entire future trajectory segment and a time index
that indicates the current execution step against the planning step.
Rather than relying only on a single waypoint, the policy observes the entire sparse
trajectory and learns to infer the proper alignment implicitly.

\subsubsection{Policy Formulation}
At  control step $t$, the policy receives
\begin{equation}
    o_t = \left[ o_t^{\mathrm{prop}},\, \mathbf{h}_t,\, \hat{i},\, o_t^{\mathrm{task}}\right],
    \label{eq:observation}
\end{equation}
and outputs PD joint targets $a_t \in \mathbb{R}^{29}$. 
The proprioceptive observation is defined as 
$o_t^{\mathrm{prop}} \triangleq
[\mathbf{q}_t,\dot{\mathbf{q}}_t,\boldsymbol{\omega}_t,\mathbf{g}_t]$,
including joint positions, joint velocities, base angular velocity, and projected gravity. 
The history buffer is defined as 
$\mathbf{h}_t \triangleq [o_{t-L_h:t-1}^{\mathrm{prop}},\mathbf{a}_{t-L_h:t-1}]$ with $L_h=25$. Given the time elapsed since the last planning step is $t-t_k$, we define time index as

\begin{equation}
    \hat{i} = \left\lfloor \frac{t-t_k}{\tau} \right\rfloor.
\end{equation}

The policy retains the entire cached trajectory and additionally receives a phase-aligned reference token, preserving both future motion context and the current execution phase. The task observation is defined as
$o_t^{\mathrm{task}} \triangleq [\mathcal{T}_k,\, \Delta \mathcal{T}_k ]$,
where  $\Delta \mathcal{T}_k$ is the local SE(3) errors between the reference and the current
end-effector and head poses at the update time $t_k$. The time index $\hat{i}$ and $o_t^{\mathrm{task}}$
jointly guide the robot to capture the asynchrony and motion context.

\subsubsection{Auxiliary Drift Estimator}
To further improve robustness under asynchronous reference-frame mismatch, we introduce an auxiliary head to estimate the base height and each end-effector's pose change relative to its pose at the latest update time $t_k$. The estimator takes the features from the observation history, and
employs a lightweight decoder for prediction.
Note that the predicted frame drift is not applied to transform the reference trajectory, but acts
as an auxiliary supervision signal for regularization.

\subsection{Teacher-Student Distillation}
\label{sec:distill}

We train the policy via teacher-student distillation,  as shown in
Fig.~\ref{fig:overview} (Stage 1). The teacher policy is pre-trained on a large-scale whole-body motion dataset with access to privileged information (e.g., globally-consistent state
signals), while the student only observes the asynchronous input defined in Sec.~\ref{sec:async}.
The loss for the student policy includes both the imitation and the
reinforcement learning:
\begin{equation}
\mathcal{L}_{\mathrm{train}}
=
\mathcal{L}_{\mathrm{imit}} + \mathcal{L}_{\mathrm{PPO}}.
\end{equation}

\subsubsection{Imitation Learning}

The imitation objective contains action distillation and auxiliary state estimation:
\begin{equation}
\mathcal{L}_{\mathrm{imit}}
=
\lambda_{\mathrm{act}}\mathcal{L}_{\mathrm{act}}
+
\lambda_{\mathrm{se}}\mathcal{L}_{\mathrm{se}} .
\end{equation}
The action distillation term is defined as
\begin{equation}
\mathcal{L}_{\mathrm{act}}
=
\left\|
\mathbf{a}^{\mathrm{stu}}_t
-
\mathbf{a}^{\mathrm{tea}}_t
\right\|_2^2 ,
\end{equation}
where $\mathbf{a}^{\mathrm{tea}}_t$ and $\mathbf{a}^{\mathrm{stu}}_t$ are the teacher and student
actions, respectively.

The auxiliary estimation loss is
\begin{equation}
\mathcal{L}_{\mathrm{se}}
=
\mathcal{L}_{\mathrm{ee}}
(\Delta\hat{\mathbf p}_t,\Delta\hat{\mathbf R}_t,
 \Delta\mathbf p_t,\Delta\mathbf R_t)
+
\mathcal{L}_{\mathrm{height}}
\left(
\hat{h}_t,h_t
\right),
\end{equation}
where, for each end effector $e\in\{\mathrm{lh},\mathrm{rh},\mathrm{head}\}$,
\[
\Delta\mathbf p_t^e=(\mathbf R_{t_k}^e)^\top(\mathbf p_t^e-\mathbf p_{t_k}^e),
\quad
\Delta\mathbf R_t^e=(\mathbf R_{t_k}^e)^\top\mathbf R_t^e.
\]
Here, $(\mathbf p_t^e,\mathbf R_t^e)$ is the executed EE pose in a common world frame; its relative change is expressed in the EE frame at $t_k$. We supervise the relative position, 6D rotation representation, and current base height using simulator ground truth available only during training.

\subsubsection{Reinforcement Learning} Since the teacher policy has access to the global observation
and reference, its action guidance may cause the ambiguity for the student. So we employ the
reinforcement learning for student policy.

The PPO loss $\mathcal{L}_{\mathrm{PPO}}$ is optimized with motion regularization and tracking
reward. Importantly, we design a sliding-window global tracking reward to improve global position
tracking within a re-planning interval.
At time $t$, we evaluate tracking in the latest planning-time base frame $b_{t_k}$, as shown in Fig.~\ref{fig:global_rwd}. Specifically, both the reference and executed end-effector poses are expressed in $b_{t_k}$, and the $SE(3)$ tracking error is
\begin{equation}
\operatorname{Log}
\left(
\left(
{}^{b_{t_k}}\mathbf{T}^{\mathrm{ref}}_{ee}(t)
\right)^{-1}
{}^{b_{t_k}}\mathbf{T}_{ee_t}
\right),
\label{eq_globalrwd}
\end{equation}
The reference at $t$ is interpolated at the controller rate. Here, $\operatorname{Log}(\cdot)$ maps the relative transform to a 6D error
in $\mathfrak{se}(3)$. This reward is evaluated at every control step and resets when a new planner
command arrives, thus penalizing drift accumulated within each planning window.

\section{\textbf{ASYNC-CA}: Completion-augmented for Post-training}
\label{sec:ASYNC-CA}

ASYNC-3PT tracks sparse head-and-hand task-space references without explicit frame alignment. 
However, direct RL post-training of this controller is not effective. 
This is because the task-specific data contain only sparse trajectories, making the standard teacher-student distillation 
pipeline inapplicable. To overcome this, we propose an MPC-based completion module to complete floating-base and upper-body joint trajectories from the sparse reference, and augment the ASYNC-3PT observation and reward, resulting in the controller ASYNC-CA. We pretrain ASYNC-CA ($\pi_0$ in Fig.~\ref{fig:overview}) using the same teacher-student distillation as Sec~\ref{sec:distill}. During post-training, the MPC-completed references provide augmented information for adaptation. Combined with two-level self-guidance, this stabilizes post-training under sparse references, as shown in Stage 2 in Fig.~\ref{fig:overview}.

\begin{table*}[t]
\centering
\caption{Comparison of different controller designs under synchronous and asynchronous execution settings}
\vspace{-0.1cm}
\label{tab:async_compare}
\begingroup
\setlength{\tabcolsep}{3pt}
\renewcommand{\arraystretch}{1.12}
\par\centering
\resizebox{\textwidth}{!}{
\begin{tabular}{lccccccccc}
    \toprule
    Method 
    & Succ. (\%) $\uparrow$
    & \makecell[c]{$\mathrm{Pos}_{1\mathrm{s}}\,\downarrow$}
    & \makecell[c]{$\mathrm{Rot}_{1\mathrm{s}}\,\downarrow$}
    & \makecell[c]{$\mathrm{Pos}_{\mathrm{loc}}\,\downarrow$}
    & \makecell[c]{$\mathrm{Rot}_{\mathrm{loc}}\,\downarrow$}
    & \makecell[c]{$\mathrm{LinVel}_{1\mathrm{s}}\,\downarrow$}
    & \makecell[c]{$\mathrm{AngVel}_{1\mathrm{s}}\,\downarrow$}
    & \makecell[c]{$\mathrm{LinVel}_{\mathrm{loc}}\,\downarrow$}
    & \makecell[c]{$\mathrm{AngVel}_{\mathrm{loc}}\,\downarrow$} \\
    \midrule
    
    \gray{TEACHER} & \gray{99.32} & \gray{/} & \gray{/} & \gray{$4.08_{\pm 1.38}$} & \gray{$10.06_{\pm 2.82}$} & \gray{/} & \gray{/} & \gray{$0.45_{\pm 0.17}$} & \gray{$0.93_{\pm 0.17}$} \\

    \addlinespace[2pt]
    \gray{Sync W.SC} & \gray{98.93} & \gray{/} & \gray{/} & \gray{$4.25_{\pm 1.46}$} & \gray{$10.43_{\pm 2.79}$} & \gray{/} & \gray{/} & \gray{$0.46_{\pm 0.17}$} & \gray{$0.90_{\pm 0.18}$} \\

    \addlinespace[2pt]
    Sync W.ASC & 89.94 & $7.86_{\pm 3.89}$ & $13.79_{\pm 5.31}$ & $6.16_{\pm 2.62}$ & $12.37_{\pm 4.05}$ & $0.53_{\pm 0.32}$ & $0.99_{\pm 0.38}$ & $0.51_{\pm 0.19}$ & $1.07_{\pm 0.29}$ \\

    \addlinespace[2pt]
    Sync W.explicit estimator & 62.16 & $6.55_{\pm 1.56}$ & $\mathbf{9.15}_{\pm 2.58}$ & $6.39_{\pm 1.67}$ & $\mathbf{9.37}_{\pm 2.80}$ & $\mathbf{0.47}_{\pm 0.14}$ & $0.80_{\pm 0.14}$ & $\mathbf{0.47}_{\pm 0.13}$ & $0.99_{\pm 0.23}$ \\

    \addlinespace[2pt]
    ASYNC-3PT from scratch & 73.63 & $12.46_{\pm 3.49}$ & $29.44_{\pm 6.79}$ & $11.51_{\pm 3.04}$ & $27.85_{\pm 6.96}$ & $0.61_{\pm 0.24}$ & $1.46_{\pm 0.41}$ & $0.62_{\pm 0.24}$ & $1.54_{\pm 0.40}$ \\

    \addlinespace[2pt]
    ASYNC-3PT (ours) & \textbf{98.93} & $\mathbf{5.91}_{\pm 1.53}$ & $11.53_{\pm 2.88}$ & $\mathbf{4.55}_{\pm 1.48}$ & $10.82_{\pm 2.96}$ & $0.49_{\pm 0.15}$ & $\mathbf{0.74}_{\pm 0.10}$ & $0.47_{\pm 0.15}$ & $\mathbf{0.90}_{\pm 0.19}$ \\

    \midrule
    Decoupled & 94.336 & $11.804_{\pm 3.733}$ & $17.678_{\pm 6.331}$ & $\mathbf{3.223}_{\pm 1.870}$ & $11.649_{\pm 5.353}$ & $0.497_{\pm 0.192}$ & $1.920_{\pm 0.917}$ & $\mathbf{0.416}_{\pm 0.188}$ & $1.676_{\pm 0.871}$ \\

    \addlinespace[2pt]
    ASYNC-CA w.o. estimator & 98.87 & $5.54_{\pm 1.74}$ & $11.36_{\pm 2.90}$ & $4.35_{\pm 1.58}$ & $10.56_{\pm 3.02}$ & $0.48_{\pm 0.18}$ & $\mathbf{0.72}_{\pm 0.11}$ & $0.47_{\pm 0.17}$ & $\mathbf{0.87}_{\pm 0.21}$ \\

    \addlinespace[2pt]
    ASYNC-CA (ours) & \textbf{99.17} & $\mathbf{4.87}_{\pm 1.46}$ & $\mathbf{10.94}_{\pm 2.77}$ & $4.08_{\pm 1.44}$ & $\mathbf{10.15}_{\pm 2.91}$ & $\mathbf{0.47}_{\pm 0.17}$ & $0.76_{\pm 0.12}$ & $0.46_{\pm 0.16}$ & $0.90_{\pm 0.21}$ \\

    \bottomrule
\end{tabular}}
\par\smallskip
\vspace{-0.3cm}
\endgroup
\end{table*}

\subsection{MPC-based Trajectory Completion and Policy Augmentation}
\label{sec:mpc}
The high-level planner provides sparse task-space references $\mathcal{T}_k$ only for the head and two hands, leaving the floating-base motion and upper-body joint configurations unspecified. We therefore use a kinematic MPC to complete these sparse references into constraint-aware base and upper-body trajectories, which serve as dense guidance for policy execution and post-training. 
We extend the original reference trajectory in~(\ref{reftraj}) with the MPC-completed floating-base trajectory and augment $o_t^{\mathrm{task}}$ in~(\ref{eq:observation}) with the completed upper-body joint references. We further introduce upper-body tracking rewards and an asynchronous sliding-window reward for base tracking. Together, these observation and reward augmentations provide denser supervision for post-training under sparse planner commands.

\subsection{Self-guidance Reinforcement Learning}
\label{sec:selfdistill}

In post-training, we use the MPC-completed base and upper-body references. These completed references provide additional upper-body and base guidance. However, the lower body is still not directly specified by the sparse task-space reference. Therefore, direct RL adaptation may satisfy the sparse tracking objective while producing abnormal lower-body motions. To preserve the pretrained motion prior, we introduce two-level self-guidance regularization during post-training, as indicated by the blue arrows in Fig.~\ref{fig:overview}(right).

We use the pretrained policy $\pi_0$ as a pseudo-teacher for the adapted policy $\pi_1$. Let $\mathbf{a}_t^1=\pi_1(\mathbf{o}_t)$ and $\mathbf{a}_t^0=\pi_0(\mathbf{o}_t)$ denote their predicted actions. The action-level self-guidance is defined as
\begin{equation} 
\mathcal{L}_{\mathrm{sd}}^{\mathrm{act}} = \mathbb{E}\left[ \left\| \mathbf{a}_t^1-\mathbf{a}_t^0 \right\|_2^2 \right]. 
\end{equation} 
This term constrains the adapted policy to stay close to the pretrained action prior, reducing jerky action variations.
We further apply FK-level self-guidance on selected body poses:
\begin{equation} 
\mathcal{L}_{\mathrm{sd}}^{\mathrm{fk}} = \mathbb{E}\left[ \left\| \mathrm{FK}(q^{\mathrm{tar}}(\mathbf{a}_t^1)) - \mathrm{FK}(q^{\mathrm{tar}}(\mathbf{a}_t^0)) \right\|_2^2 \right], 
\end{equation} 
where $\mathrm{FK}(\cdot)$ maps the decoded target joint configuration to selected body poses. This term regularizes the adapted motion in Cartesian space. In practice, we upweight lower-body poses in the FK-level loss because they are not directly constrained by sparse head-and-hand tracking and are more prone to drift.

Together, the dense rewards from MPC-completed references and the two-level self-guidance regularization allow the policy to adapt to novel tasks while avoiding catastrophic drift from the pretrained motion prior.



\section{Experimental Results}
We evaluate the proposed framework along two axes. First, we study whether asynchronous modeling is necessary for sparse trajectory tracking under low-frequency references, and analyze tracking performance across different update frequencies.
Second, we examine whether MPC-completed guidance improves post-training beyond sparse 3-point guidance on out-of-distribution tasks, and evaluate the effect of self-guidance in both simulation and real-robot experiments.

 \begin{figure*}
    \centering
    \includegraphics[width=\linewidth]{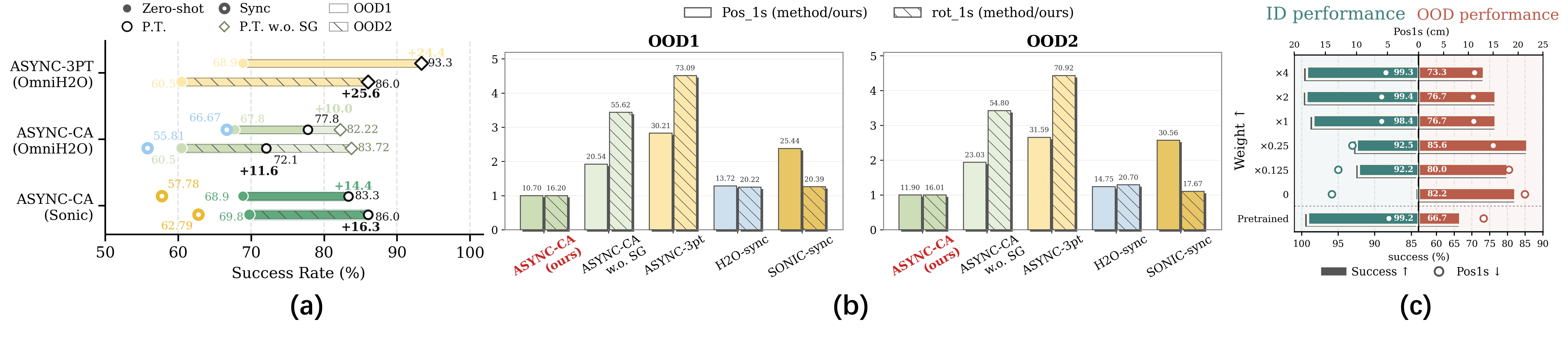}
      \vspace{-0.8cm}
    \caption{
\textbf{OOD Adaptation under Sparse Task-Space References.}
(a) Success rates on \textit{ood1} and held-out \textit{ood2}. Filled markers denote zero-shot deployment and hollow markers denote post-training (P.T.). Half-filled markers denote the original synchronous policy under asynchronous updates, which serves as a baseline. ASYNC-3PT uses only three-point end-effector references; ASYNC-CA uses MPC-completed base and upper-body guidance with self-guidance.
(b) Interval tracking errors on \textit{ood1} and \textit{ood2}. Each bar reports the ratio of $\mathrm{Pos}_{1\mathrm{s}}$ or $\mathrm{Rot}_{1\mathrm{s}}$ to ASYNC-CA (ours), where ASYNC-CA is normalized to $1.0$; larger values indicate larger asynchronous drift.
(c)  Comparison of different self-guidance weights. Bars show the absolute success rate and circles show $\mathrm{Pos}_{1\mathrm{s}}$.
}
    \label{fig:ood_test1} 
    \vspace{-0.2cm}
 \end{figure*}
\subsection{Experimental Setup and Metrics}

We evaluate the proposed framework on the 29-DoF Unitree G1 humanoid in simulation and on the real robot. 
The controller receives sparse future trajectories with a horizon of 20 frames and runs at 50~Hz. 
Unless otherwise specified, the sparse references are updated at 1~Hz, corresponding to a 1~s reference interval. 

We report success rate and tracking errors.
Following SONIC~\cite{sonic}, \textit{Succ.} counts a rollout as successful if it completes without falling or exceeding a vertical tracking error of $0.25$~m at any tracked end effector.
Our primary metrics are the asynchronous interval errors: \textbf{$\mathrm{Pos}_{1\mathrm{s}}$} and \textbf{$\mathrm{Rot}_{1\mathrm{s}}$}, which are computed over one asynchronous reference interval and measure frame drift from the latest planning time using the same planning-frame error definition as Eq.~\eqref{eq_globalrwd}. The local metrics \textbf{$\mathrm{Pos}_{\mathrm{loc}}$} and \textbf{$\mathrm{Rot}_{\mathrm{loc}}$} are computed in the root-relative frame and measure local end-effector tracking accuracy. All position errors are reported in centimeters, rotation errors in degrees, linear velocity errors in m/s, and angular velocity errors in rad/s.
Table~\ref{tab:async_compare} reports statistics over all rollouts, including failures.

\subsection{Effect of Asynchronous Sparse Trajectory Tracking}

We first evaluate whether explicit asynchronous modeling is necessary for executing low-frequency sparse references. We consider the following controller variants:
\begin{itemize}
    \item \textbf{\textsc{TEACHER}:} a privileged teacher policy adopted from the OmniH2O framework~\cite{omnih2o}.
    
    \item \textbf{Sync W.SC:} a synchronous sparse tracker trained in the OmniH2O frame~\cite{omnih2o}, evaluated with synchronized references and the same future horizon.
    
    \item \textbf{Sync W.ASC:} the same synchronous tracker is directly deployed under asynchronous reference updates, distilled from \textsc{TEACHER}.

    \item \textbf{ASYNC-3PT from scratch:} ASYNC-3PT trained directly from scratch with RL. This baseline examines whether sparse head-and-hand references alone are sufficient for asynchronous tracking.
    \item \textbf{Sync W.explicit estimator:} a synchronous tracker that explicitly corrects cached references using learned drift estimates.
    
\end{itemize}
Since both \textsc{TEACHER} and Sync W.SC use unavailable information, they are reported only as references. The remaining controllers are evaluated under asynchronous reference updates. All methods are trained and evaluated using the motion dataset from OMOMO and GRAB~\cite{omomo,grab}.

\textbf{Effect of asynchronous modeling.} As shown in Table~\ref{tab:async_compare}, ASYNC-3PT achieves $98.93\%$ success and reduces the interval position error from $7.86$~cm for Sync W.ASC to $5.91$~cm, demonstrating more reliable asynchronous execution.

\begin{center}
\vspace{-0.01cm}
\begin{minipage}{\columnwidth}
\captionof{table}{\normalfont Ablation of time-indexed conditioning on a held-out trajectory set.}
\label{tab:time_idx_ablation}
\vspace{-0.15cm}
\centering
\resizebox{\columnwidth}{!}{
\begin{tabular}{cccccc}
\toprule
Time index & Succ. $\uparrow$ & Pos$_{1s}$ $\downarrow$ & Rot$_{1s}$ $\downarrow$ & Vel$_{1s}$ $\downarrow$ & AngVel$_{1s}$ $\downarrow$ \\
\midrule
\xmark & 98.9 & 14.6 & 26.6 & 0.550 & 1.298 \\
\cmark & \textbf{99.4} & \textbf{8.6}  & \textbf{12.8} & \textbf{0.416} & \textbf{1.048} \\
\bottomrule
\end{tabular}}
\end{minipage}
\end{center}


\textbf{Effect of time-indexed conditioning.}
We further ablate the time-indexed conditioning on a separate unseen trajectory set. 
As shown in Table~\ref{tab:time_idx_ablation}, removing the time index only slightly affects success rate, but increases $\mathrm{Pos}_{1\mathrm{s}}$ from 8.6~cm to 14.6~cm and $\mathrm{Rot}_{1\mathrm{s}}$ from 12.8$^\circ$ to 26.6$^\circ$. 
This confirms that the time index mainly helps resolve the phase mismatch within each asynchronous reference interval.
The time index also enables inference-time speed modulation, while higher re-planning frequencies reduce tracking errors~\cite{supplementary_appendix}.

\begin{figure*}
    \centering
    \includegraphics[width=\linewidth]{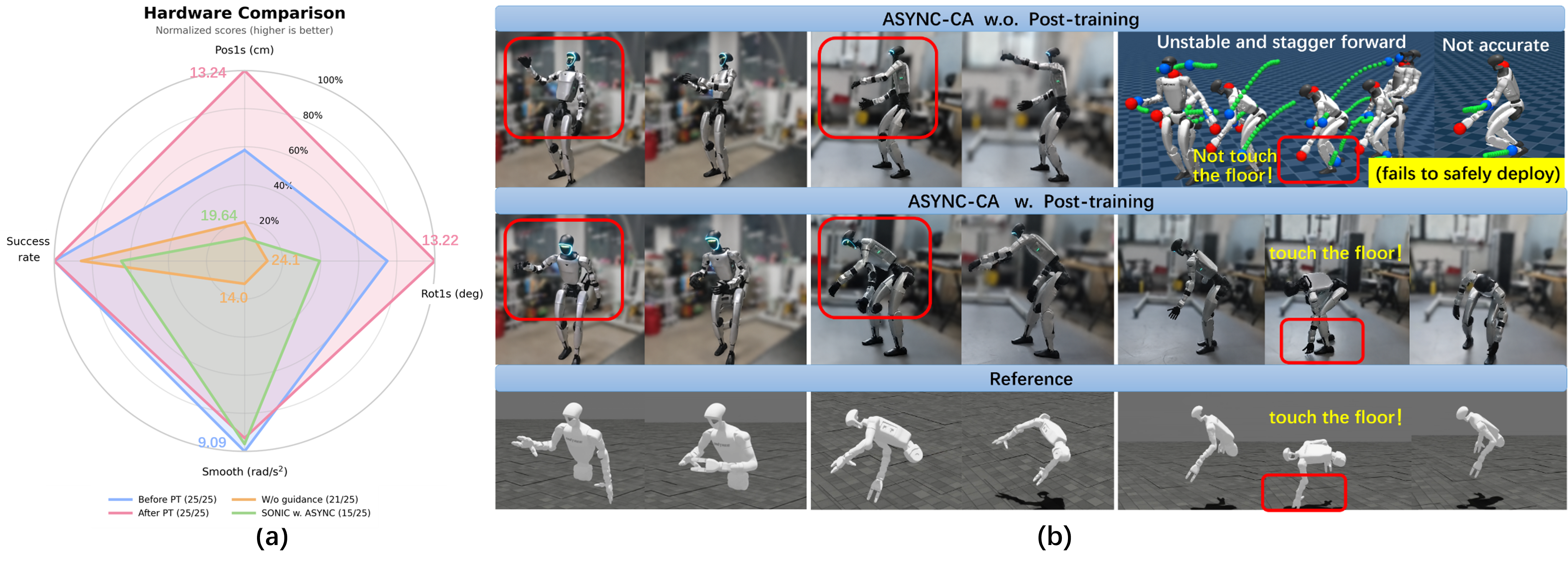}
      \vspace{-0.8cm}
    \caption{Hardware comparison and qualitative post-training evaluation.
    (a) Five-motion hardware comparison (25 trials per policy). Each metric is reverse min--max normalized across the four policies (higher is better); radial percentages denote normalized scores, and legend counts denote successful trials.
    (b)Without post-training, \textbf{ASYNC-CA} exhibits inaccurate tracking and unstable forward staggering, where the reference motion is nearly in-place. With post-training, the policy better follows the reference and maintains a stable whole-body posture in both simulation and real-world execution. Red boxes highlight representative differences.
    } 
    \vspace{-0.3cm}
    \label{fig:post-train-real}
\end{figure*}

\subsection{Comparison of Base-augmented Tracking}

We compare the last two controller variants in Table~\ref{tab:async_compare}.
The Decoupled baseline uses the same base and upper-body references and observations as ASYNC-CA, but separates base tracking and upper-body tracking into two different control modules.
This design is related to recent decoupled tracking frameworks such as HoldMyBeer~\cite{holdmybeer} and PMP~\cite{PMP}.
Although Decoupled achieves lower local position error, ASYNC-CA reduces asynchronous errors. Thus, better root-relative accuracy does not imply better asynchronous tracking.

We also ablate the state estimator in ASYNC-CA.
Removing the auxiliary estimator slightly reduces success and increases interval position/rotation errors. This indicates a modest pose-tracking benefit from auxiliary supervision, without requiring its predictions for online reference correction.


\subsection{Closed-loop Online Manipulation}
\label{sec:online_planning}

\begin{figure}[H]
\centering
\vspace{-0.2cm}
\includegraphics[width=0.9\linewidth]{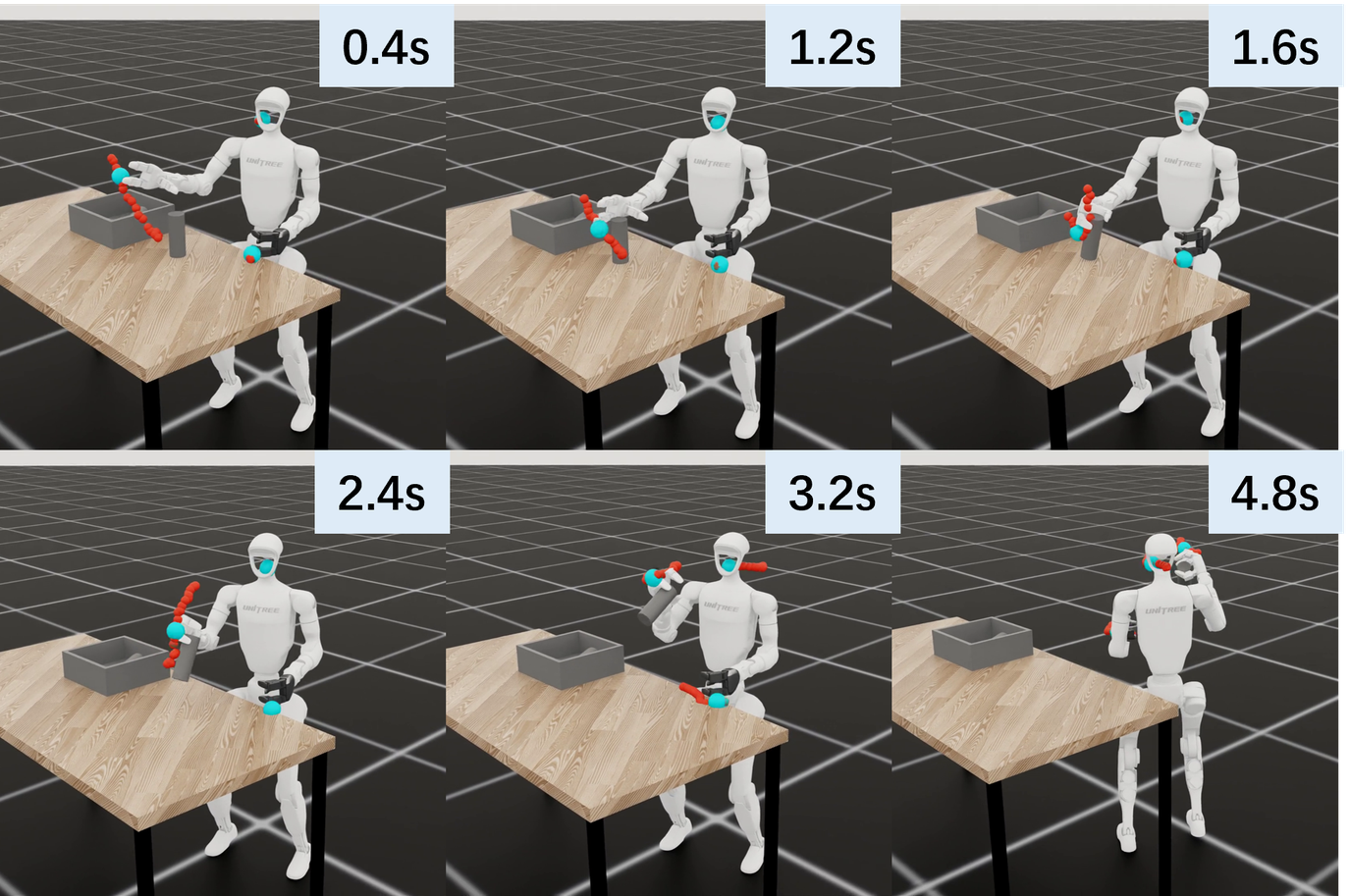}
\vspace{-0.2cm}
\caption{GR00T+ASYNC-CA cup-grasping execution with online planner.}
\vspace{-0.4cm}
\label{fig:online_planning_main}
\end{figure}

\begin{table}[H]
\centering
\caption{Online closed-loop task comparison.}
\vspace{-0.1cm}
\label{tab:online_planning_main}
\setlength{\tabcolsep}{3pt}
\par\centering\resizebox{\columnwidth}{!}{%
\begin{tabular}{llcc}
\toprule
Method & Planner interface & Success $\uparrow$ & Tracking error $\downarrow$ \\
\midrule
\makecell[l]{GR00T +\\ASYNC-CA} & \textbf{Sparse:} 3-EE traj. & \textbf{12/25} & $\mathbf{6.69}_{\pm 0.11}$ / $\mathbf{8.41}_{\pm 0.58}$ \\
$\Psi_0$ + AMO & \makecell[l]{\textbf{Dense:} base vel. +\\upper-body joints} & 6/25 & $7.23_{\pm 0.44}$ / $9.29_{\pm 0.67}$ \\
\bottomrule
\vspace{-0.5cm}
\end{tabular}}\par
\end{table}

To measure closed-loop tracking errors with a real-time planner, we integrate ASYNC-CA with a GR00T-based planner~\cite{grootn1}, which predicts head-and-hand trajectories from the current observation every 0.8~s. Online MPC completes them into base and upper-body references. We also train a $\Psi_0$+AMO baseline that predicts dense upper-body and base action chunks. Both planners use the same 200 simulated demonstrations collected with VIRAL~\cite{VIRAL} and are evaluated on 25 cup-grasping trials from random initial positions.
Table~\ref{tab:online_planning_main} reports success rates and asynchronous position and rotation tracking errors. Our pipeline achieves higher success (48\% vs.\ 24\%) and lower tracking errors, while requiring the planner to predict only three sparse end-effector trajectories rather than dense base-and-upper-body action chunks. Since the two planners use different interfaces and horizons, this comparison evaluates the complete online pipelines rather than isolating the controller alone. Under identical $\Psi_0$ predictions at 0.46-s intervals, ASYNC-CA also yields lower and more stable tracking errors than AMO~\cite{supplementary_appendix}.

\subsection{OOD Post-Training with MPC Completion}
\label{sec:ood_posttrain}

We construct a failure-mined hard-motion OOD benchmark from AMASS by zero-shot evaluating the OmniH2O teacher and selecting 120 trajectories on which it fails or incurs large tracking errors. These trajectories do not overlap with the pretraining or standard evaluation data. We use 80 motions for post-training (\emph{ood1}) and hold out 40 for evaluation (\emph{ood2}). Using OmniH2O~\cite{omnih2o} and SONIC~\cite{sonic} as privileged teacher backbones, we compare ASYNC-3PT, post-trained via RL alone on the original sparse head-and-hand references, with ASYNC-CA, post-trained using MPC-completed base and upper-body guidance with and without self-guidance (Fig.~\ref{fig:ood_test1}).

\textbf{Effect of MPC-completed guidance.}
RL-only post-training of ASYNC-3PT achieves higher success under the OmniH2O backbone but leaves large asynchronous tracking errors, especially rotational drift. Combining MPC-completed base and upper-body guidance with self-guidance, ASYNC-CA reduces the \emph{ood1} position/rotation errors from $30.21$~cm/$73.09^\circ$ to $10.70$~cm/$16.20^\circ$, with similar reductions on held-out \emph{ood2} (Fig.~\ref{fig:ood_test1}(b)). After post-training, ASYNC-CA also achieves higher success and lower interval errors than synchronous controllers deployed under asynchronous reference updates (Fig.~\ref{fig:ood_test1}(a,b)), supporting the effectiveness of the combined guidance for OOD adaptation.

\textbf{Effect of self-guidance.}
We jointly scale the action- and FK-level self-guidance weights by $w\in\{0,0.125,0.25,1,2,4\}$ and evaluate ID retention on OMOMO+GRAB and adaptation on \emph{ood1}.
Figure~\ref{fig:ood_test1}(c) reveals a clear adaptation--preservation trade-off. Without self-guidance, post-training improves \emph{ood1} success by 15.5 percentage points, but ID success drops from 99.2\% to 84.5\%, while the OOD position error increases to 21.36~cm. In contrast, $w=2$ preserves ID success while still improving OOD success by 10.0 percentage points and reducing the position error to 10.96~cm. Overall, stronger self-guidance better preserves the pretrained ID behavior, whereas weaker guidance permits greater OOD adaptation at the cost of increased policy drift.

\textbf{Effect of teacher backbone.}
With the SONIC teacher, ASYNC-CA reaches 83.3\% success on \emph{ood1} and 86.0\% on \emph{ood2}. The consistent improvement obtained with the same post-training procedure indicates that the method is not tied to a particular teacher backbone.

\subsection{Real-world Hardware Experiments }
\label{sec:real_world_hardware}

\textbf{Quantitative hardware evaluation.} We deploy 14 motions spanning walking, bending/squatting, kneeling, standing up, and manipulation on G1, with $60/70$ trials completed without falling (Table~\ref{tab:motion_hardware}). On the five motions selected by zero-shot simulation success, P.T. improves tracking while retaining $25/25$ success (Fig.~\ref{fig:post-train-real}(a)). Smooth is the temporal mean of the joint-wise RMS acceleration of 17 measured upper-body joints (lower is better). Removing self-guidance reduces success to $21/25$ and increases Smooth from 9.476 to 14.008~rad/s$^2$, showing that self-guidance improves both robustness and motion smoothness.

\begin{table}[t]
\centering
\caption{
Simulation and hardware performance on\\ 14 selected motions for real-world deployment. Simulation metrics use all-motion statistics.
}
\label{tab:motion_hardware}
\small
\setlength{\tabcolsep}{3.5pt}
\renewcommand{\arraystretch}{1.08}
\resizebox{\columnwidth}{!}{%
\begin{tabular}{@{}llcccc@{}}
\toprule
Eval. & Policy & Succ. & $\mathrm{Pos}_{1s}\downarrow$ (cm) & $\mathrm{Rot}_{1s}\downarrow$ ($^\circ$) & $\mathrm{Smooth}\downarrow$ (rad/s$^2$) \\
\midrule
\multirow{4}{*}{Simulation} & Zero-shot & 5/14 (35.7\%) & 12.455 & 13.270 & 13.35 \\
& w/o SG & 6/14 (42.9\%) & 12.727 & 13.592 & 14.465 \\
& SONIC-w. ASY & 6/14 (42.9\%) & 28.095 & 18.129 & 11.810 \\
& \textbf{After PT} & \textbf{12/14 (85.7\%)} & \textbf{11.273} & \textbf{12.147} & \textbf{11.50} \\
\midrule
Hardware & \textbf{After PT} & \textbf{60/70 (85.7\%)} & 13.42 & 14.32 & 10.52 \\
\bottomrule
\vspace{-0.5cm}
\end{tabular}}
\end{table}

\begin{figure}[!t]
  \vspace{0.18cm}
  \centering
  \includegraphics[width=\linewidth]{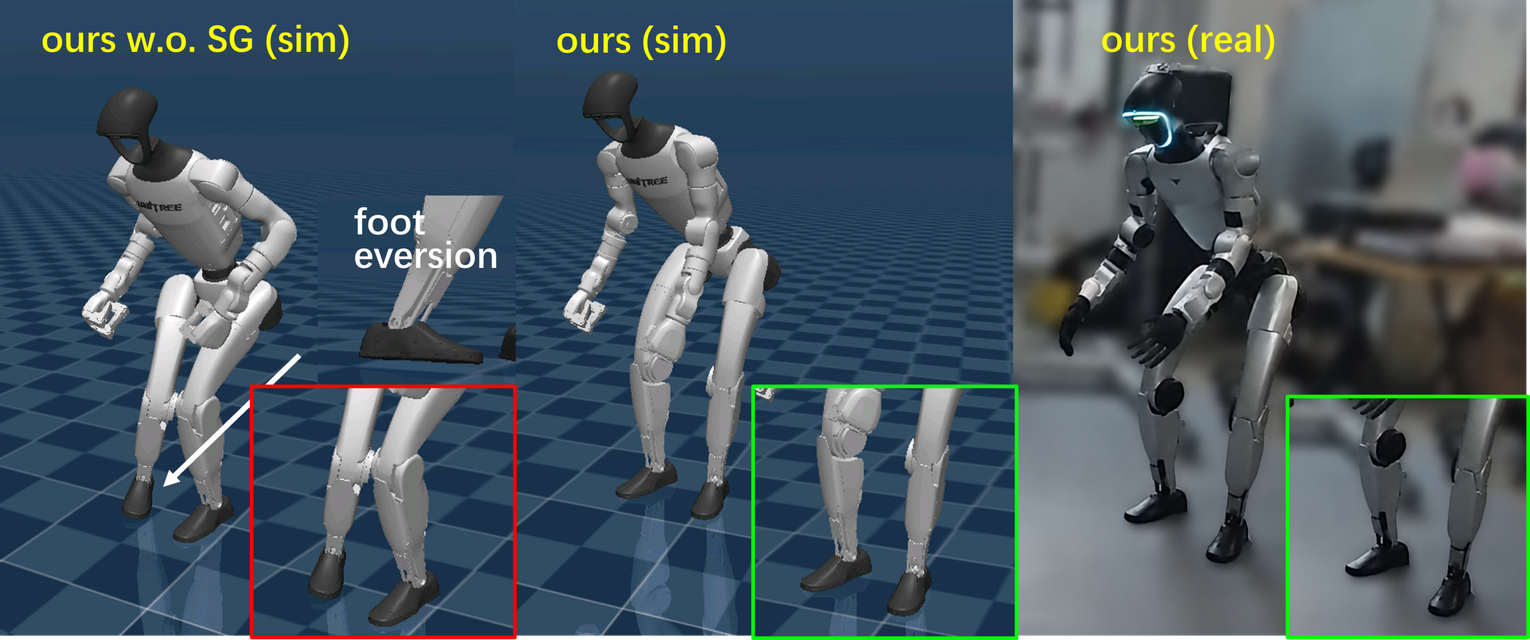}
  \vspace{-0.4cm}
  \caption{We compare post-trained policies with and without self-guidance(SG). Removing self-guidance leads to unrealistic lower-body motions, such as foot eversion. Self-guidance regularizes the lower body and improves real-world transfer.}
  \vspace{-0.2cm}
  \label{fig:no-sg}
\end{figure}

\textbf{Qualitative self-guidance evaluation.} Without self-guidance, the underconstrained lower body develops unrealistic motions such as foot eversion; self-guidance preserves natural lower-body behavior and improves real-world transfer (Fig.~\ref{fig:no-sg}).

\section{Conclusion}
This letter presented an asynchronous upper-body task-space trajectory tracking policy for humanoid robots, using time‑indexed conditioning and sliding‑window rewards for  handling of asynchronous planning-frame mismatch, without online explicit reference reprojection, plus MPC‑based completion and self‑guidance for stable RL adaptation. Simulation and hardware validation on the G1 confirms effectiveness and improvements. Future work may investigate the end-to-end training of both planner and controller.


\bibliographystyle{IEEEtran}
\bibliography{IEEEabrv,ref}

\clearpage
\appendices

\section{MPC Formulation}
\label{sec:mpc_appendix}

We use finite-horizon kinematic MPC to complete three end-effector (EE) pose/twist tracks into a virtual floating-base and 17-joint upper-body trajectory for G1. The optimization uses forward kinematics, without torque, contact-force, inertia, or rigid-body acceleration variables.

\subsection{State, Input, and Reference Representation}

The state and generalized-velocity input are
\begin{equation}
\begin{aligned}
\mathbf{x}&=[\mathbf{p}_b^\top,\boldsymbol{\Theta}_b^\top,
\mathbf{q}^{\mathrm{up},\top}]^\top\in\mathbb{R}^{23},\\
\mathbf{u}&=[\dot{\mathbf{p}}_b^\top,\dot{\boldsymbol{\Theta}}_b^\top,
\dot{\mathbf{q}}^{\mathrm{up},\top}]^\top\in\mathbb{R}^{23}.
\end{aligned}
\end{equation}
Here, $\mathbf{p}_b=[p_x,p_y,p_z]^\top$ and $\boldsymbol{\Theta}_b=[\psi,\theta,\phi]^\top$ describe the virtual base using ZYX yaw--pitch--roll coordinates. The 17 joints are ordered as waist yaw/roll/pitch, followed by the left and right arms (shoulder pitch/roll/yaw, elbow, wrist roll/pitch/yaw). At update $t_k$, the prediction model is
\begin{equation}
\dot{\mathbf{x}}(t)=\mathbf{u}(t),\qquad
\mathbf{x}(t_k)=\mathbf{x}^{\mathrm{obs}}_k.
\end{equation}
Forward kinematics maps the generalized coordinates to the end-effector poses $\mathbf{T}_e(\mathbf{x})=[\mathbf{R}_e(\mathbf{x}),\mathbf{p}_e(\mathbf{x})]$, where $e\in\mathcal{E}=\{\mathrm{lh},\mathrm{rh},\mathrm{head}\}$ denotes the left hand, right hand, and head.

The reference contains timestamped EE poses and twists, together with state/input references:
\begin{equation}
\mathcal{R}_k=\{(t_j,\bar{\mathbf{T}}_{e,j},\bar{\mathbf{v}}_{e,j},
\bar{\boldsymbol{\omega}}_{e,j},\bar{\mathbf{x}}_j,\bar{\mathbf{u}}_j)\}_{j=0}^{N_{\mathrm{ref}}-1}.
\end{equation}
EE poses and the observed state use the same planning-time frame $b_{t_k}$. Reference positions are linearly interpolated, and orientations use spherical linear interpolation of unit quaternions. Missing twists are estimated from adjacent samples, e.g., $\bar{\mathbf{v}}_{e,j}\simeq(\bar{\mathbf{p}}_{e,j+1}-\bar{\mathbf{p}}_{e,j})/(t_{j+1}-t_j)$. These are references, not future state measurements.

The prediction horizon is $T_{\mathrm{mpc}}=2.0$~s, with multiple-shooting interval $\Delta t=0.02$~s. The reference sequence contains $N_{\mathrm{ref}}=20$ waypoints at $\tau=0.1$~s spacing, spanning 0--1.9~s; reference sampling and solver discretization are distinct. The base-height reference is $0.75$~m and the yaw reference follows the head orientation within $[-\pi,\pi]$; pitch and roll use the supplied base reference or zero. Base $x,y$ remain optimization variables with zero direct state-tracking weight.

\subsection{Tracking Objective and Soft Constraints}

The running cost is
\begin{equation}
\begin{aligned}
\ell(\mathbf{x},\mathbf{u},t)
&=\tfrac12\|\mathbf{x}-\bar{\mathbf{x}}(t)\|_Q^2
+\tfrac12\|\mathbf{u}-\bar{\mathbf{u}}(t)\|_R^2\\
&\quad+\sum_{e\in\mathcal{E}}\ell_e^{\mathrm{EE}}
+\ell_{\mathrm{lim}}+\ell_{\mathrm{vel}}+\ell_{\mathrm{col}},
\end{aligned}
\end{equation}
where $\|\mathbf{a}\|_W^2=\mathbf{a}^\top W\mathbf{a}$. The state weights are $Q=\operatorname{diag}(Q_b,Q_q)$, with
\begin{equation}
\begin{aligned}
Q_b&=\operatorname{diag}(0,0,0.5,5,0.5,1),\\
Q_q&=\operatorname{diag}(1,1,1,Q_a,Q_a),\\
Q_a&=\operatorname{diag}(0.6,0.4,0.6,0.2,0.3,0.3,0.3).
\end{aligned}
\end{equation}
The input reference is zero, and the diagonal matrix $R$ weights base and joint velocities separately. The state reference $\bar{\mathbf{x}}$ specifies the posture prior independently of the EE targets.

Writing each quaternion as $\mathbf{q}_e=(\mathbf{q}_{v,e},q_{w,e})$, define
\begin{equation}
\begin{aligned}
\mathbf{r}_e^p&=\begin{bmatrix}\mathbf{p}_e(\mathbf{x})-\bar{\mathbf{p}}_e\\
\mathbf{e}_{R,e}\end{bmatrix},\\
\mathbf{e}_{R,e}&=q_{w,e}\bar{\mathbf{q}}_{v,e}-\bar q_{w,e}\mathbf{q}_{v,e}
+\mathbf{q}_{v,e}\times\bar{\mathbf{q}}_{v,e},\\
\mathbf{r}_e^v&=\begin{bmatrix}\bar{\mathbf{v}}_e-\mathbf{v}_e(\mathbf{x},\mathbf{u})\\
\bar{\boldsymbol{\omega}}_e-\boldsymbol{\omega}_e(\mathbf{x},\mathbf{u})\end{bmatrix}.
\end{aligned}
\end{equation}
The EE penalty is
\begin{equation}
\ell_e^{\mathrm{EE}}=\tfrac12\|\mathbf{r}_e^p\|_{W_{p,e}}^2
+\tfrac12\|\mathbf{r}_e^v\|_{W_{v,e}}^2.
\end{equation}
Both matrices have the form $\operatorname{diag}(w_1I_3,w_2I_3)$. Pose position/orientation weights are $(40000,9600)$ for either hand and $(16000,2400)$ for the head; twist linear/angular weights are $(25,10)$ and $(10,5)$, respectively. EE tracking is soft, without hard pose equalities.

For an inequality $h\geq0$, the relaxed log barrier is
\begin{equation}
B_{\mu,\delta}(h)=
\begin{cases}
-\mu\ln h,&h>\delta,\\[1mm]
\mu\left[-\ln\delta+\frac12\left(\left(\frac{h-2\delta}{\delta}\right)^2-1\right)\right],&h\leq\delta.
\end{cases}
\end{equation}
The costs $\ell_{\mathrm{lim}}$ and $\ell_{\mathrm{vel}}$ sum these barriers over the 17 joint-position bounds and all 23 input-velocity bounds, with $(\mu,\delta)=(0.1,10^{-4})$ and $(0.01,10^{-3})$, respectively. For the selected pairs of robot links $c$,
\begin{equation}
\ell_{\mathrm{col}}=\sum_c B_{0.2,10^{-3}}(d_c(\mathbf{x})-d_{\min}),
\qquad d_{\min}=0.08~\mathrm{m}.
\end{equation}
The quadratic extension below $\delta$ gives finite violation costs, so these terms do not provide hard feasibility guarantees.

\subsection{Optimal Control Problem and MPC Output}

At each update, the MPC solves
\begin{equation}
\begin{aligned}
\min_{\mathbf{x}(\cdot),\mathbf{u}(\cdot)}\;&
\Phi(\mathbf{x}(t_k+T_{\mathrm{mpc}}))
+\int_{t_k}^{t_k+T_{\mathrm{mpc}}}\ell(\mathbf{x},\mathbf{u},t)\,dt\\
\text{s.t.}\;&\dot{\mathbf{x}}=\mathbf{u},\qquad
\mathbf{x}(t_k)=\mathbf{x}^{\mathrm{obs}}_k.
\end{aligned}
\end{equation}
We use no terminal cost ($\Phi=0$); position, velocity, and collision limits are represented by the soft penalties above. The solution comprises the optimized state and input trajectories $(\mathbf{x}^*,\mathbf{u}^*)$. The first six state components give base translation and ZYX orientation, and the last 17 give upper-body joint angles; the input follows the same partition for generalized velocities.

The completed base and upper-body trajectory is sampled at 20 waypoints and included in the policy observation. Phase-aligned interpolation provides auxiliary tracking rewards, and the reference sequence is retained until the next 1-Hz MPC update. The MPC resets every 1.0~s and initializes each new solve from the latest observation. Offline replay and online operation use the same receding-horizon formulation with different reference sources.

\subsection{Solver, Initialization, and Runtime}

We solve the optimal control problem using multiple-shooting sequential quadratic programming (SQP) in OCS2, with Euler discretization of the dynamics and HPIPM for the quadratic subproblems. The 2.0-s horizon has 100 intervals and 101 state nodes. The baseline and higher-accuracy settings use at most two and eight SQP iterations, with two and eight computational threads, respectively. The SQP step and cost tolerances are $10^{-4}$. Each quadratic subproblem uses at most 15 iterations, with stationarity, equality, inequality, and complementarity tolerances of $(10^{-4},10^{-6},10^{-6},10^{-6})$, respectively. Only the optimized trajectories are required for reference completion.

The measured state fixes the initial condition. The nominal input is initialized from its reference (zero for offline trajectories), and the nominal state follows $\dot{\mathbf{x}}=\mathbf{u}$. Offline initialization may additionally use the first-frame base pose and joint angles. During online operation, the initial EE and upper-body joint references are anchored to the observation at the corresponding time and smoothly blended into the incoming trajectory to improve continuity. The previous solution may warm-start subsequent optimizations; after a reset, initialization proceeds from the current observation.

Runtime evaluation on ten OMOMO-GRAB trajectories used two outer SQP iterations and eight CPU threads on an AMD Ryzen Threadripper PRO 7995WX: pure solution time was $36.44\pm3.97$~ms (range $29.87$--$41.63$~ms). These timings cover the two-iteration optimization only, excluding reference preparation and trajectory communication.

\subsection{Scope and Limitations}

The completion stage maps three EE pose/twist tracks to a 23-dimensional base-plus-upper-body reference through forward kinematics and numerical optimization, without a separate inverse-kinematics, retargeting, or learned pose-completion stage. The completed trajectory is kinematic guidance; its soft costs alone do not guarantee dynamic whole-body stability, collision freedom, or real-time execution.

\section{OOD Generalization and MPC-induced Reference Shift}
\label{sec:MPC-induced-ood}

\begin{figure}[!t]
\centering
\includegraphics[width=0.88\linewidth]{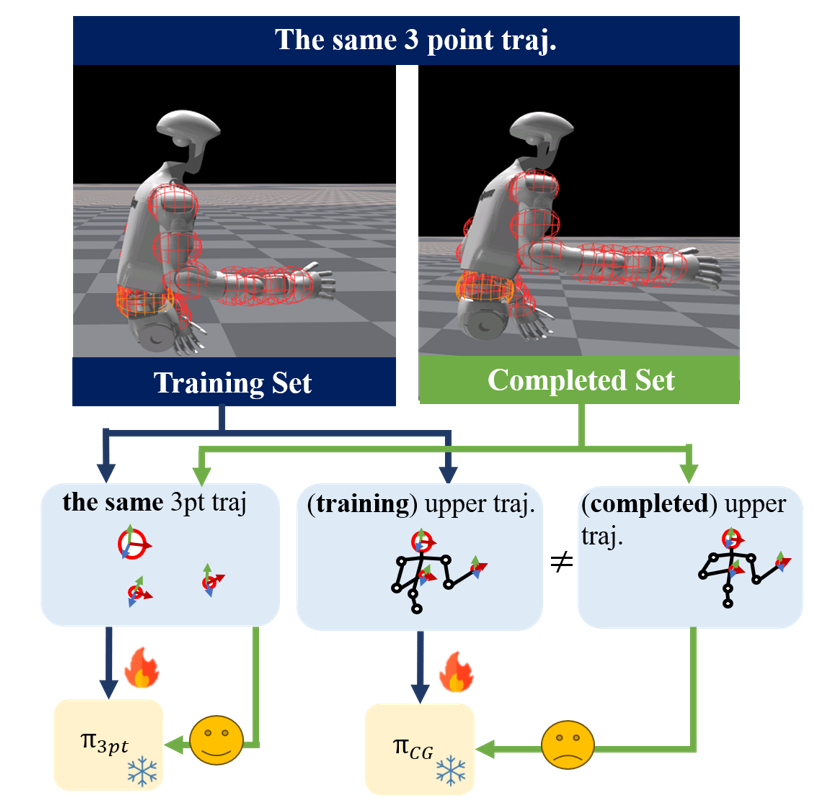}
\caption{MPC-induced reference shift under zero-shot deployment: original and completed references for the same sparse three-point trajectory.}
\label{fig:reason_for_ood}
\end{figure}

Figure~\ref{fig:reason_for_ood} illustrates the different base and upper-body references completed from the same sparse three-point trajectory. Table~\ref{tab:ood_eval} quantifies the resulting policy-input shift on the same 20 motions. The first two conditions evaluate the unchanged pretrained policy with original and MPC-completed references. The post-trained conditions use references consistent with training; post-training with original references serves as a ground-truth completion oracle.

\begin{table*}[!t]
\centering
\caption{\normalfont Policy evaluation with original and MPC-completed references on the same 20 motions.}
\label{tab:ood_eval}
\resizebox{\textwidth}{!}{
\begin{tabular}{lllcccc}
\toprule
Policy & Reference & Statistics & Succ. (\%) $\uparrow$ & Pos$_{1\mathrm{s}}$ (cm) $\downarrow$ & Rot$_{1\mathrm{s}}$ ($^\circ$) $\downarrow$ & Margin$_{p05}$ $\uparrow$ \\
\midrule
\multirow{2}{*}{Original} & \multirow{2}{*}{Original} & Successful & \textbf{65} & \textbf{$\bm{23.42\pm2.86}$} & \textbf{$\bm{22.51\pm2.55}$} & \textbf{$\bm{-0.01501}$} \\
 & & All & & \textbf{$\bm{23.94\pm3.75}$} & \textbf{$\bm{23.56\pm4.62}$} & \textbf{$\bm{-0.01578}$} \\
\midrule
\multirow{2}{*}{Original} & \multirow{2}{*}{MPC-completed} & Successful & 55 & $25.06\pm4.96$ & $25.39\pm7.03$ & $-0.01529$ \\
 & & All & & $23.99\pm5.37$ & $26.18\pm5.81$ & $-0.01752$ \\
\midrule
\multirow{2}{*}{Post-trained} & \multirow{2}{*}{MPC-completed} & Successful & 90 & $12.66\pm4.64$ & $21.35\pm9.02$ & \textbf{$\bm{-0.00620}$} \\
 & & All & & $12.90\pm4.46$ & $21.93\pm8.82$ & \textbf{$\bm{-0.00549}$} \\
\midrule
Post-trained & Original & Successful = All & \textbf{100} & \textbf{$\bm{10.96\pm2.97}$} & \textbf{$\bm{18.13\pm4.92}$} & $-0.00974$ \\
\bottomrule
\end{tabular}}
\par\smallskip
\begin{minipage}{\textwidth}
\footnotesize\raggedright
$\mathrm{Margin}_{p05}$ is the frame-averaged 5th percentile across joints of $m_{t,j}=\min(q_{t,j}-l_j,u_j-q_{t,j})/(u_j-l_j)$, where $[l_j,u_j]$ is the central 85\% of the URDF joint range. Positive values retain soft-limit clearance; negative values indicate soft-limit, but not necessarily physical-limit, violations.
Mean frame-wise MPC-to-original position differences for the left hand, right hand, and head are 2.778/2.798/1.968~mm (2.514~mm overall); the corresponding mean per-motion maxima are 11.388/12.527/8.632~mm.
\end{minipage}
\end{table*}

With the unchanged pretrained policy, replacing original references with MPC-completed references reduces success from 65\% to 55\%. Post-training on MPC-completed references raises success to 90\% and lowers the all-rollout position/rotation errors from $23.99$~cm/$26.18^\circ$ to $12.90$~cm/$21.93^\circ$. The original-reference oracle reaches 100\% success and $10.96$~cm/$18.13^\circ$, quantifying the remaining completion-induced gap. Thus, MPC preserves the sparse three-point task at millimeter scale, while its generated base and upper-body references create a policy-input shift that post-training substantially mitigates.

\section{Controller-Rate Phase Index}
\label{app:time_index}

Unless otherwise specified, each cached reference contains $N_{\mathrm{ref}}=20$ future waypoints at $\tau=0.1$~s intervals, indexed from 0 to 19 and spanning 0--1.9~s within the nominal 2.0-s horizon. The controller period is $\Delta t_c=0.02$~s, and the planner update interval is at most 1~s and set to $\Delta t_k=1$~s during training.

The phase index $\hat{i}$ in Sec.~\ref{sec:async} is retained as the elapsed
waypoint-bin index. In implementation, the elapsed controller-step count is
\begin{equation}
\hat{i}_c =
\left\lfloor
\frac{t-t_k}{\Delta t_c}
\right\rfloor .
\end{equation}
It resets to zero whenever a new reference
arrives and ranges from 0 to 49 within a 1-s update interval. If a deployed reference carries its generation
timestamp $t_k$ but arrives at $a_k \geq t_k$, the phase can instead be
initialized as $\hat{i}_c(a_k)=\left\lfloor(a_k-t_k)/\Delta t_c\right\rfloor$,
rather than zero. Since
$\tau=0.1$~s$=5\Delta t_c$, the waypoint-bin index is:
\begin{equation}
\hat{i} =
\left\lfloor
\frac{\hat{i}_c}{5}
\right\rfloor .
\end{equation}
For access to the zero-based cached waypoint sequence $\{\mathbf{T}_{k,i}\}_{i=0}^{N_{\mathrm{ref}}-1}$, the corresponding index is
\begin{equation}
i_t =
\operatorname{clip}
\left(
\hat{i},\,
0,\,
N_{\mathrm{ref}}-1
\right).
\end{equation}
The policy observes $\hat{i}_c$ and the complete cached trajectory rather than using $i_t$ alone to select an instantaneous target. For reward computation, the continuous phase
$\operatorname{clip}(\hat{i}_c\Delta t_c/\tau,0,N_{\mathrm{ref}}-1)$ is used to interpolate
the reference as described in Appendix~\ref{app:reward}.

\subsection{Inference-Time Modulation and Re-planning Frequency}

\begin{figure}[!t]
  \centering
  \includegraphics[width=0.75\linewidth]{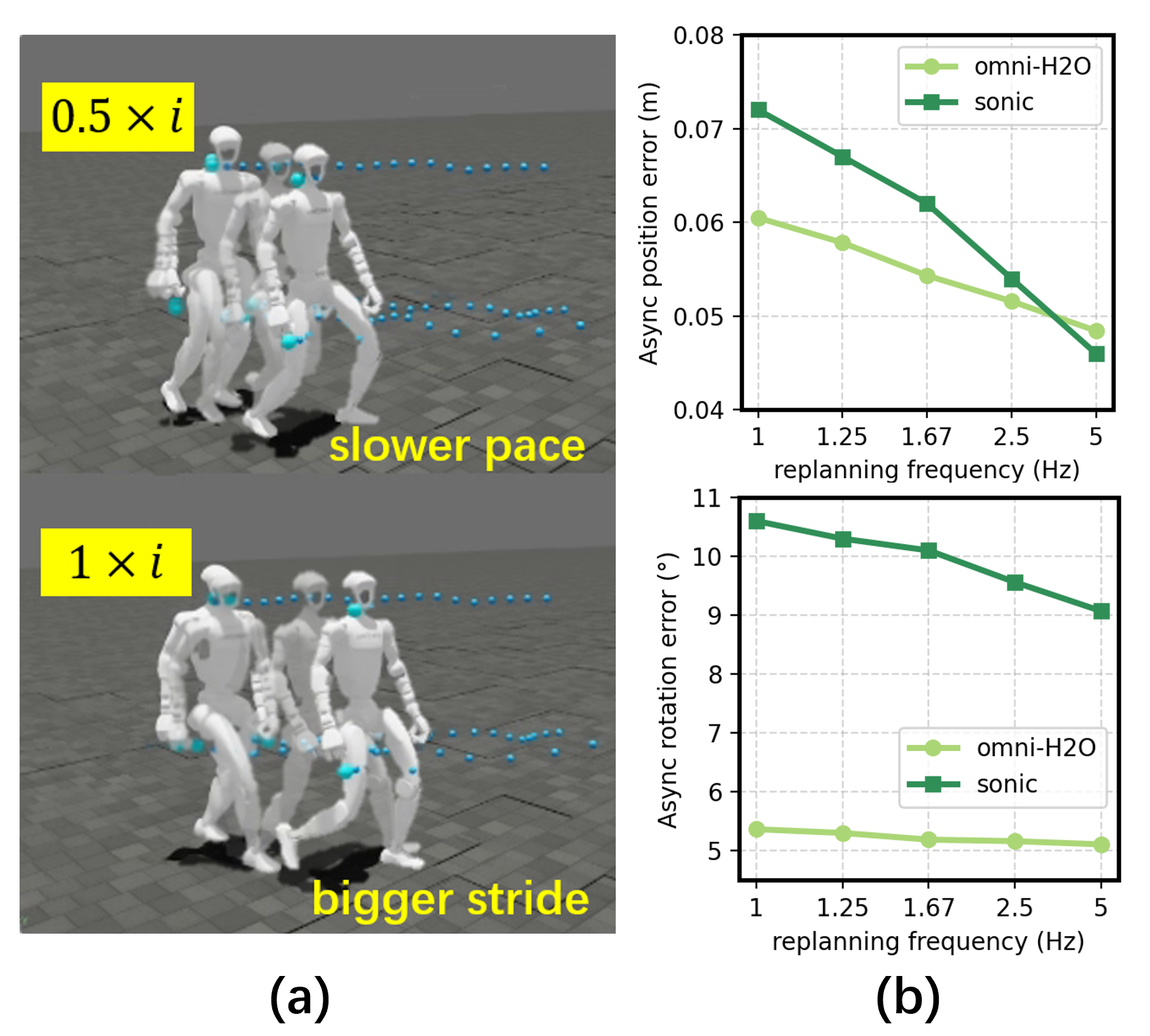}
  \vspace{-0.3cm}
  \caption{(a) Inference-time speed modulation enabled by time-indexed conditioning. (b) Asynchronous tracking error under different re-planning frequencies with OmniH2O- and SONIC-based teachers.}
  \vspace{-0.3cm}
  \label{fig:replan_freq}
\end{figure}

Beyond tracking, the time index also enables inference-time speed modulation. By changing its progression rate, the same reference can be executed faster or slower. As shown in Fig.~\ref{fig:replan_freq}(a), a slower progression leads to slower execution with shorter strides, while the original progression produces faster motion with larger strides.

\textbf{Effect of re-planning frequency.} We evaluate asynchronous tracking across different re-planning frequencies. As shown in Fig.~\ref{fig:replan_freq}(b), our method remains stable for all tested rates and is not tied to a specific controller frequency. Higher re-planning frequencies reduce position and rotation errors because shorter execution intervals alleviate frame mismatch and error accumulation. The same trend appears with both OmniH2O- and SONIC-based~\cite{sonic} teachers, suggesting that this frequency effect is not specific to a particular teacher backbone.

\section{Pseudo-global Three-Keypoint Tracking Reward}
\label{app:reward}

This section details the implementation of the sliding-window global tracking reward
introduced in Sec.~IV-B. While Eq.~(11) formulates the tracking error in a unified
SE(3) form, the actual reward decomposes position and rotation into separate
exponential kernels with independent temperature parameters, allowing finer control
over the two error scales.

\paragraph{Reference interpolation.}
At control step $t$, let $n_t$ be the number of elapsed control steps since the
last reference update. The continuous reference index is
$u_t = \mathrm{clip}(n_t \Delta t \,\nu,\; 0,\; T{-}1)$,
where $\Delta t$ is the control timestep, $\nu$ the trajectory sampling rate,
and $T$ the cached horizon length. The reference position and orientation of
keypoint $i$ are obtained by linear and spherical-linear interpolation between
the two adjacent cached waypoints $\lfloor u_t \rfloor$ and
$\min(\lfloor u_t \rfloor{+}1,\,T{-}1)$ with fractional part
$\alpha = u_t - \lfloor u_t \rfloor$, yielding
$\mathbf{p}^{r}_i(t)$ and $\mathbf{q}^{r}_i(t)$.
Accordingly, the tracking error is evaluated at every 50~Hz control
step rather than only at the 0.1~s waypoints.

Let $\mathcal{K}=\{\mathrm{LH},\mathrm{RH},\mathrm{Head}\}$ denote the three
task-relevant keypoints. At each asynchronous reference update, the cached future
trajectory is expressed in the planning-time base frame $b_{t_k}$, which remains
fixed until the next update. Between two consecutive updates, both the reference
and executed keypoint poses are evaluated in this same frame.

\paragraph{Reward formulation.}
The position and rotation errors are
\begin{align}
  e_p(t) &= \frac{1}{|\mathcal{K}|}
    \sum_{i\in\mathcal{K}}
    \left\|\mathbf{p}^{r}_i(t)-\mathbf{p}^{c}_i(t)\right\|_2^2 , \label{eq:ep}\\
  e_R(t) &= \frac{1}{|\mathcal{K}|}
    \sum_{i\in\mathcal{K}}
    \left|\mathrm{angle}\!\left(
      \mathbf{q}^{r}_i(t)\otimes\mathbf{q}^{c}_i(t)^{-1}
    \right)\right| , \label{eq:er}
\end{align}
where $\mathbf{p}^{r}_i(t),\mathbf{q}^{r}_i(t)$ and
$\mathbf{p}^{c}_i(t),\mathbf{q}^{c}_i(t)$ are the reference and current keypoint
poses, respectively, all expressed in the planning-time base frame $b_{t_k}$.
The per-step reward is
\begin{equation}
  r_{\mathrm{3pt}}(t) = \Delta t \left(
    w_p \exp\!\left(-\frac{e_p(t)}{\sigma_p}\right)
    + w_R \exp\!\left(-\frac{e_R(t)}{\sigma_R}\right)
  \right),
  \label{eq:r3pt}
\end{equation}
For ASYNC-3PT, the local and pseudo-global position/orientation weights are both $(80,80)$. For ASYNC-CA, the local pair is $(10,10)$ and the pseudo-global pair is $(60,60)$, with additional base-position, base-orientation, and upper-body joint-reference weights of 10, 10, and 5. The temperatures are $\sigma_p = 0.05$ and $\sigma_R = 0.3$\,rad. The factor
$\Delta t$ normalizes the reward magnitude with respect to the control
frequency. This reward is active only when the asynchronous reference cache is
available; otherwise it returns zero.

\begin{table*}[!t]
\centering
\caption{Complete controller comparison with paired success-only and all-rollout statistics.}
\vspace{-0.1cm}
\label{tab:async_compare_full}
\begingroup
\setlength{\tabcolsep}{3pt}
\renewcommand{\arraystretch}{1.12}
\par\centering
\resizebox{\textwidth}{!}{
\begin{tabular}{lccccccccc}
    \toprule
    Method 
    & Succ. (\%) $\uparrow$
    & \makecell[c]{$\mathrm{Pos}_{1\mathrm{s}}\,\downarrow$}
    & \makecell[c]{$\mathrm{Rot}_{1\mathrm{s}}\,\downarrow$}
    & \makecell[c]{$\mathrm{Pos}_{\mathrm{loc}}\,\downarrow$}
    & \makecell[c]{$\mathrm{Rot}_{\mathrm{loc}}\,\downarrow$}
    & \makecell[c]{$\mathrm{LinVel}_{1\mathrm{s}}\,\downarrow$}
    & \makecell[c]{$\mathrm{AngVel}_{1\mathrm{s}}\,\downarrow$}
    & \makecell[c]{$\mathrm{LinVel}_{\mathrm{loc}}\,\downarrow$}
    & \makecell[c]{$\mathrm{AngVel}_{\mathrm{loc}}\,\downarrow$} \\
    \midrule
    
    \gray{TEACHER} & \gray{99.32} & \gray{/} & \gray{/} & \gray{$4.04_{\pm 1.16}$} & \gray{$9.98_{\pm 2.35}$} & \gray{/} & \gray{/} & \gray{$0.44_{\pm 0.04}$} & \gray{$0.86_{\pm 0.13}$} \\
    & & \gray{/} & \gray{/} & \gray{$4.08_{\pm 1.38}$} & \gray{$10.06_{\pm 2.82}$} & \gray{/} & \gray{/} & \gray{$0.45_{\pm 0.17}$} & \gray{$0.93_{\pm 0.17}$} \\

    \addlinespace[2pt]
    \gray{Sync W.SC} & \gray{98.93} & \gray{/} & \gray{/} & \gray{$4.20_{\pm 1.35}$} & \gray{$10.36_{\pm 2.56}$} & \gray{/} & \gray{/} & \gray{$0.46_{\pm 0.04}$} & \gray{$0.89_{\pm 0.18}$} \\
    & & \gray{/} & \gray{/} & \gray{$4.25_{\pm 1.46}$} & \gray{$10.43_{\pm 2.79}$} & \gray{/} & \gray{/} & \gray{$0.46_{\pm 0.17}$} & \gray{$0.90_{\pm 0.18}$} \\
    \addlinespace[2pt]
    Sync W.ASC & 89.94 & $6.95_{\pm 2.38}$ & $12.74_{\pm 3.23}$ & $5.59_{\pm 1.76}$ & $11.63_{\pm 2.87}$ & $0.49_{\pm 0.07}$ & $0.90_{\pm 0.15}$ & $0.48_{\pm 0.05}$ & $1.01_{\pm 0.19}$ \\
    & & $7.86_{\pm 3.89}$ & $13.79_{\pm 5.31}$ & $6.16_{\pm 2.62}$ & $12.37_{\pm 4.05}$ & $0.53_{\pm 0.32}$ & $0.99_{\pm 0.38}$ & $0.51_{\pm 0.19}$ & $1.07_{\pm 0.29}$ \\

    \addlinespace[2pt]
    ASYNC-3PT from scratch & 73.63 & $11.68_{\pm 2.54}$ & $27.96_{\pm 5.77}$ & $10.97_{\pm 2.56}$ & $26.40_{\pm 5.84}$ & $0.55_{\pm 0.10}$ & $1.33_{\pm 0.24}$ & $0.56_{\pm 0.07}$ & $1.43_{\pm 0.22}$ \\
    & & $12.46_{\pm 3.49}$ & $29.44_{\pm 6.79}$ & $11.51_{\pm 3.04}$ & $27.85_{\pm 6.96}$ & $0.61_{\pm 0.24}$ & $1.46_{\pm 0.41}$ & $0.62_{\pm 0.24}$ & $1.54_{\pm 0.40}$ \\

    \addlinespace[2pt]
    Sync W.explicit estimator & 62.16 & $6.64_{\pm 1.46}$ & $\mathbf{9.21}_{\pm 2.00}$ & $6.46_{\pm 1.57}$ & $\mathbf{9.40}_{\pm 2.16}$ & $\mathbf{0.47}_{\pm 0.07}$ & $0.78_{\pm 0.08}$ & $\mathbf{0.46}_{\pm 0.05}$ & $1.00_{\pm 0.19}$ \\
    & & $6.55_{\pm 1.56}$ & $\mathbf{9.15}_{\pm 2.58}$ & $6.39_{\pm 1.67}$ & $\mathbf{9.37}_{\pm 2.80}$ & $\mathbf{0.47}_{\pm 0.14}$ & $0.80_{\pm 0.14}$ & $0.47_{\pm 0.13}$ & $0.99_{\pm 0.23}$ \\
    \addlinespace[2pt]
    ASYNC-3PT (ours) & \textbf{98.93} & $\mathbf{5.85}_{\pm 1.37}$ & $11.45_{\pm 2.67}$ & $\mathbf{4.49}_{\pm 1.28}$ & $10.71_{\pm 2.52}$ & $0.48_{\pm 0.07}$ & $\mathbf{0.74}_{\pm 0.09}$ & $0.47_{\pm 0.04}$ & $\mathbf{0.90}_{\pm 0.18}$ \\
    &  & $\mathbf{5.91}_{\pm 1.53}$ & $11.53_{\pm 2.88}$ & $\mathbf{4.55}_{\pm 1.48}$ & $10.82_{\pm 2.96}$ & $0.49_{\pm 0.15}$ & $\mathbf{0.74}_{\pm 0.10}$ & $0.47_{\pm 0.15}$ & $\mathbf{0.90}_{\pm 0.19}$ \\
    
    \midrule
    Decoupled & 94.34 & $11.43_{\pm 3.25}$ & $16.99_{\pm 3.04}$ & $\mathbf{2.99}_{\pm 0.70}$ & $11.00_{\pm 2.02}$ & $0.47_{\pm 0.09}$ & $1.80_{\pm 0.40}$ & $\mathbf{0.39}_{\pm 0.09}$ & $1.56_{\pm 0.37}$ \\
    & & $11.80_{\pm 3.73}$ & $17.68_{\pm 6.33}$ & $\mathbf{3.22}_{\pm 1.87}$ & $11.65_{\pm 5.35}$ & $0.50_{\pm 0.19}$ & $1.92_{\pm 0.92}$ & $\mathbf{0.42}_{\pm 0.19}$ & $1.68_{\pm 0.87}$ \\

    \addlinespace[2pt]
    ASYNC-CA w.o. estimator & 98.87 & $5.47_{\pm 1.58}$ & $11.28_{\pm 2.67}$ & $4.28_{\pm 1.38}$ & $10.44_{\pm 2.51}$ & $0.47_{\pm 0.07}$ & $\mathbf{0.72}_{\pm 0.09}$ & $0.46_{\pm 0.04}$ & $\mathbf{0.87}_{\pm 0.18}$ \\
    & & $5.54_{\pm 1.74}$ & $11.36_{\pm 2.90}$ & $4.35_{\pm 1.58}$ & $10.56_{\pm 3.02}$ & $0.48_{\pm 0.18}$ & $\mathbf{0.72}_{\pm 0.11}$ & $0.47_{\pm 0.17}$ & $\mathbf{0.87}_{\pm 0.21}$ \\
    
    \addlinespace[2pt]
    ASYNC-CA (ours) & \textbf{99.17} & $\mathbf{4.82}_{\pm 1.36}$ & $\mathbf{10.88}_{\pm 2.58}$ & $4.03_{\pm 1.24}$ & $\mathbf{10.06}_{\pm 2.43}$ & $\mathbf{0.46}_{\pm 0.07}$ & $0.75_{\pm 0.00}$ & $0.45_{\pm 0.04}$ & $0.90_{\pm 0.18}$ \\
    & & $\mathbf{4.87}_{\pm 1.46}$ & $\mathbf{10.94}_{\pm 2.77}$ & $4.08_{\pm 1.44}$ & $\mathbf{10.15}_{\pm 2.91}$ & $\mathbf{0.47}_{\pm 0.17}$ & $0.76_{\pm 0.12}$ & $0.46_{\pm 0.16}$ & $0.90_{\pm 0.21}$ \\

    \bottomrule
\end{tabular}}
\par\smallskip
\scriptsize First row: successful rollouts; second row: all rollouts. Bold indicates the best mean in each comparison group and row type, excluding the gray reference methods.
\endgroup
\end{table*}

\section{Student Policy and Experimental Details}
\label{app:student_architecture}
\label{app:training_evaluation}
\subsection{Architecture}
The student receives the current proprioception $\mathbf{o}^{\mathrm{prop}}_t\in\mathbb{R}^{64}$, a cached future reference $\mathbf{o}^{\mathrm{task}}_t\in\mathbb{R}^{20\times80}$, the asynchronous phase index $\hat{i}_c$, and a proprioceptive history $\mathbf{h}_t\in\mathbb{R}^{25\times93}$. A learned $80$-D embedding of $\hat{i}_c$ conditions a temporal CNN for the reference, which outputs $\mathbf{z}_{\mathrm{ref}}\in\mathbb{R}^{160}$ and retains its phase-aligned reference token. A second temporal CNN maps the history to $\mathbf{z}_{\mathrm{hist}}\in\mathbb{R}^{240}$.

An auxiliary drift branch maps the history feature, current observation, and time embedding to a $64$-D drift embedding. During training, a decoder predicts each head/hand pose change relative to its own pose at the latest update time, expressed in that update-time EE frame, together with the current base height; it is used only for auxiliary supervision. The actor concatenates the current observation, previous action, reference and history features, time embedding, aligned token, and drift embedding into a $717$-D feature, and a $(512,512,128)$ MLP outputs the mean $\boldsymbol{\mu}\in\mathbb{R}^{29}$. DAgger directly supervises $\boldsymbol{\mu}$ with the teacher action, and deployment executes this mean action.

\begin{figure}[!t]
    \centering
    \includegraphics[width=\linewidth]{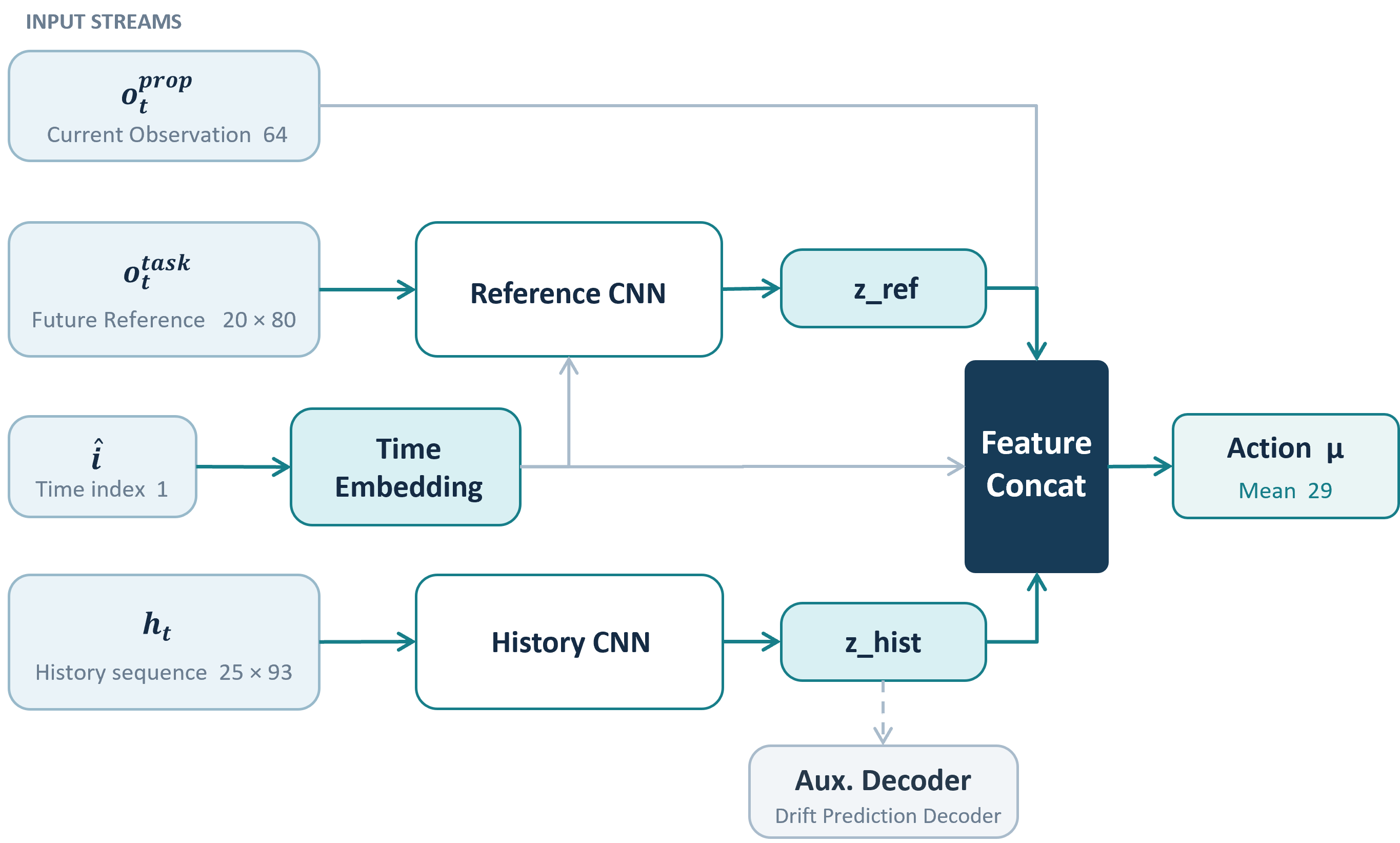}
    \vspace{-0.4em}
    \caption{Student-policy architecture. The diagram shows the ASYNC-CA configuration, whose future-reference input contains 20 frames with 80 dimensions per frame. The temporal reference and history encoders, phase-aligned reference token, and auxiliary drift embedding are fused by the actor; the drift decoder is used only during training.}
    \label{fig:student_architecture}
\end{figure}

\subsection{PPO and Control Settings}
The students jointly optimize PPO and DAgger supervision from a fixed teacher, with an action-imitation coefficient of $0.95$. Each iteration collects 24 steps in 2,048 parallel environments (49,152 transitions), followed by five optimization epochs with four minibatches per epoch. PPO uses a clipping ratio of $0.2$, discount $\gamma=0.99$, GAE $\lambda=0.95$, value-loss coefficient $1.0$, entropy coefficient $0.005$, and clipped value loss. The adaptive learning rate starts at $10^{-3}$, targets a KL divergence of $0.01$, and is bounded within $[3\times10^{-5},10^{-2}]$; the gradient-norm limit is $0.2$. The auxiliary end-effector position, orientation, and base-height losses each have unit weight.

The controller outputs 29 joint-position targets at 50~Hz, with action scale $0.25$, a $0.005$-s simulation step, and decimation of four. Unfiltered PD control uses $(K_p,K_d)=(40.179,2.558)$ for hip yaw/pitch, $(99.098,6.309)$ for hip roll/knee, $(28.501,0.907)$ for ankle pitch/roll, $(80.358,2.558)$ for the waist, and $(14.251,0.907)$ for shoulder/elbow/wrist joints.

\subsection{Domain Randomization}
Both policies use base-CoM offsets in $[-0.1,0.1]$~m, selected-link mass scaling in $[0.8,1.2]$, horizontal pushes every 5~s up to 1~m/s, RFI-limit scaling in $[0.5,1.5]$, 0--2 control-cycle delay (0--40~ms), and reference-position perturbations of $\pm2$~cm in $x/y$ and $\pm10$~cm in $z$. The static friction coefficient,
dynamic friction coefficient, and restitution coefficient are sampled from
$[0.3,1.6]$, $[0.3,1.2]$, and $[0,0.5]$, respectively. A joint zero-position
offset sampled from $\mathcal{U}(-0.01,0.01)$\,rad is applied to every joint.
Training is performed on mildly uneven generated terrain containing random
grid cells and uniform height-field perturbations with heights between
$0.001$ and $0.005$\,m. All domain-randomization terms and reset-state
perturbations are disabled during evaluation.

\subsection{Evaluation Protocol}
For Table~\ref{tab:async_compare}, each method is evaluated with 2,048 rollouts from a single seed. A rollout ends at reference completion or failure; failure occurs upon falling or when the vertical tracking error of any tracked end effector exceeds $0.25$~m.

We first compute each metric per rollout, then report its mean $\pm$ standard deviation across the 2,048 trajectories. Thus, $\pm$ denotes trajectory-level variability. Table~\ref{tab:async_compare_full} reports paired statistics: the first row for each method uses successful rollouts only; the second uses all rollouts, including failed trajectories up to and including termination. For ASYNC-3PT, the success-only subset contains 2,026 rollouts. Table~\ref{tab:async_compare} in the main text retains only the all-rollout results. Success rate is computed over all evaluation rollouts. The main-text analysis of Table~\ref{tab:async_compare} uses all-rollout errors.

\subsection{Self-guidance Sweep and Lower-body Deviation}
\label{app:self_guidance}
All variants start from the same pretrained policy, use 2,000 training epochs, and jointly scale the action- and FK-level guidance weights by $w\in\{0,0.125,0.25,1,2,4\}$; $w=1$ is the default. Table~\ref{tab:appendix_self_guidance_sweep} reports ID retention and OOD1 adaptation. Joint margin uses the soft-limit definition in Appendix~\ref{sec:MPC-induced-ood}. Smooth. is the temporal mean of the joint-wise RMS acceleration of commanded joint targets, computed by second differences at $\Delta t_c=0.02$~s and aggregated across motions.

\begin{table*}[!t]
\centering
\caption{Effect of self-guidance weight on ID retention and OOD adaptation.}
\label{tab:appendix_self_guidance_sweep}
\begin{minipage}{0.95\textwidth}
\centering
\resizebox{0.95\linewidth}{!}{%
\begin{tabular}{c|ccccc}
\hline
\multicolumn{6}{c}{\textbf{(a) OMOMO+GRAB (ID)}} \\
\hline
\textbf{Weight}
& \textbf{Succ. $\uparrow$ (\%)}
& $\mathrm{Pos}_{1\mathrm{s}}$ \textbf{$\downarrow$ (cm)}
& $\mathrm{Rot}_{1\mathrm{s}}$ \textbf{$\downarrow$ ($^\circ$)}
& \textbf{Joint margin $\uparrow$}
& \textbf{Smooth. $\downarrow$ (rad/s$^2$)} \\
\hline
Pretrained
& 99.17 & $4.82 \pm 1.36$ & $10.88 \pm 2.58$
& -0.01 & 413.16 \\

$0$
& 84.47 & $13.97 \pm 3.35$ & $48.40 \pm 6.59$
& -0.04 & 624.67 \\

$\times0.125$
& 92.19 & $12.92 \pm 5.29$ & $33.10 \pm 4.22$
& -0.03 & 677.34 \\

$\times0.25$
& 92.48 & $10.67 \pm 3.01$ & $18.12 \pm 3.28$
& -0.00 & 546.91 \\

$\times1$
& 98.44 & $5.96 \pm 2.01$ & $11.99 \pm 2.95$
& -0.01 & 441.50 \\

$\times2$
& \textbf{99.37} & $5.97 \pm 2.01$ & $11.33 \pm 2.82$
& -0.01 & \textbf{430.84} \\

$\times4$
& 99.32 & $\mathbf{5.27 \pm 1.59}$ & $\mathbf{11.11 \pm 2.81}$
& -0.01 & 435.01 \\
\hline
\end{tabular}}
\par\medskip
\resizebox{0.95\linewidth}{!}{%
\begin{tabular}{c|ccccc}
\hline
\multicolumn{6}{c}{\textbf{(b) OOD1}} \\
\hline
\textbf{Scale}
& \textbf{Succ. $\uparrow$ (\%)}
& $\mathrm{Pos}_{1\mathrm{s}}$ \textbf{$\downarrow$ (cm)}
& $\mathrm{Rot}_{1\mathrm{s}}$ \textbf{$\downarrow$ ($^\circ$)}
& \textbf{Joint margin $\uparrow$}
& \textbf{Smooth. $\downarrow$ (rad/s$^2$)} \\
\hline
Pretrained & 66.67 & $13.04 \pm 5.44$ & $16.49 \pm 6.26$ & -0.01 & 680.42 \\
$0$ & 82.22 & $21.36 \pm 7.37$ & $57.13 \pm 6.72$ & -0.04 & 545.47 \\
$\times0.125$ & 80.00 & $18.13 \pm 6.50$ & $33.54 \pm 7.30$ & -0.02 & 570.03 \\
$\times0.25$ & \textbf{85.56} & $15.00 \pm 6.23$ & $22.18 \pm 6.02$ & 0.01 & 563.03 \\
$\times1$ & 76.67 & $11.08 \pm 3.96$ & $16.57 \pm 6.27$ & -0.00377 & 604.33 \\
$\times2$ & 76.67 & $\mathbf{10.96 \pm 4.04}$ & $\mathbf{16.37 \pm 6.23}$ & -0.00639 & 644.82 \\
$\times4$ & 73.33 & $11.21 \pm 4.21$ & $16.56 \pm 6.17$ & -0.00684 & 679.74 \\
\hline
\end{tabular}}
\par\smallskip
This sweep is conducted as an independent experiment, with the same newly sampled seed set.
\end{minipage}
\end{table*}

\begin{table}[!t]
\centering
\caption{Lower-body FK deviation from $\pi_0$, computed from decoded joint targets.}
\label{tab:appendix_lower_body_fk}
\resizebox{\columnwidth}{!}{%
\begin{tabular}{llcc}
\hline
\noalign{\vskip 2pt}
\multicolumn{1}{c}{\multirow{2}{*}{\textbf{Guidance}}} & \multicolumn{1}{c}{\multirow{2}{*}{\textbf{Eval.}}} & \textbf{FK-pos (cm)} & \textbf{FK-rot} \\
& & \textbf{deviation} & \textbf{deviation} \\
\noalign{\vskip 2pt}
\hline
no self-guidance & succ\ only & 3.919 & $11.584^\circ$ \\
                 & all          & 3.929 & $11.608^\circ$ \\
$\times1$ self-guidance & succ\ only & 2.513 & $6.898^\circ$ \\
                       & all          & \textbf{2.52}  & $\mathbf{6.915^\circ}$ \\
\hline
\end{tabular}}
\end{table}
Self-guidance preserves ID performance while allowing OOD adaptation. On OOD1, default guidance reduces all-rollout lower-body FK position/rotation deviations from $3.929$~cm/$11.608^\circ$ to $2.52$~cm/$6.915^\circ$ (Table~\ref{tab:appendix_lower_body_fk}), supporting reduced drift from the pretrained motion prior.

\raggedbottom
\section{Online VLA Comparison}
\label{app:vla_comparison}
\begin{figure*}[!t]
\centering
\setlength{\abovecaptionskip}{0.15cm}
\includegraphics[width=\textwidth]{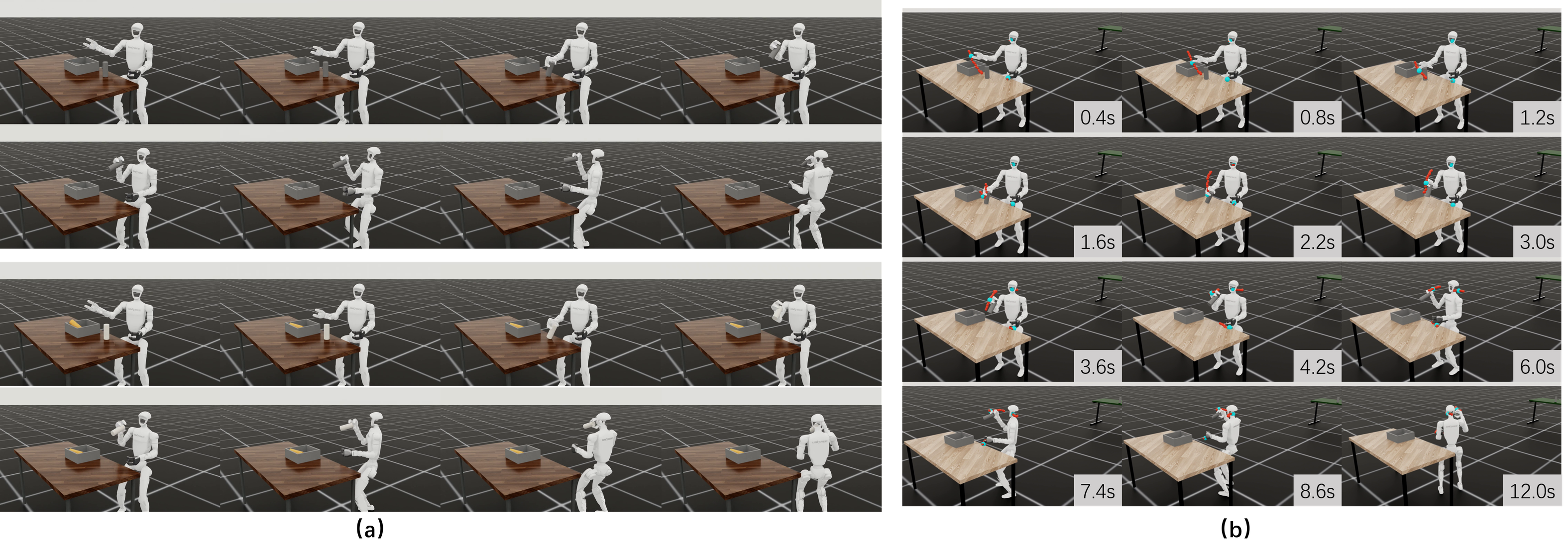}
\caption{GR00T+ASYNC-CA evaluation. Left: two representative cup-grasping executions. Right: online execution with the predicted references and their timestamps.}
\label{fig:vla_online_grasping}
\end{figure*}

We further evaluate ASYNC-CA with a GR00T-based online planner~\cite{grootn1} on cup grasping. Every $0.8$~s, GR00T predicts sparse Cartesian trajectories for the head and both hands, and online MPC completes the base and upper-body references. We compare this pipeline with $\Psi_0$~\cite{psi0}+AMO~\cite{amo}, which predicts $0.46$-s action chunks containing base velocities and upper-body joint configurations. Both planners are trained with the same 200 simulated demonstrations and evaluated over 25 trials from randomly sampled initial positions. For evaluation, the $\Psi_0$ outputs are rolled out through forward kinematics, and both pipelines use the same planning-frame $SE(3)$ error definition. Representative GR00T+ASYNC-CA executions are shown in Fig.~\ref{fig:vla_online_grasping}.

\begin{figure}[H]
    \centering
    \includegraphics[width=0.9\linewidth]{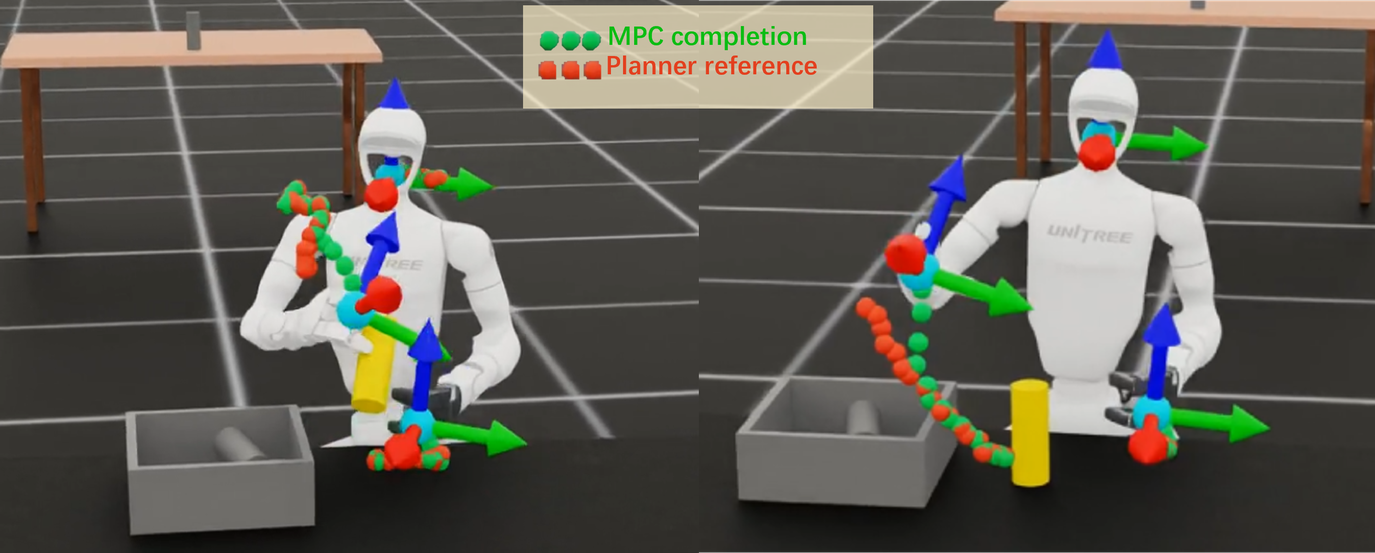}
    \caption{Online MPC completion from the robot's current observed configuration. The MPC-completed trajectories (green) smoothly connect the current head-and-hand poses to the newly received planner references (red).}
    \label{fig:r3_mpc_online_smoothing}
\end{figure}

Online MPC anchors each horizon to the observed configuration, $\mathbf{x}(0)=\mathbf{x}^{\mathrm{obs}}_k$.  Figure~\ref{fig:r3_mpc_online_smoothing} illustrates the smooth transition to newly received references.

\begin{figure*}[!t]
\centering
\setlength{\abovecaptionskip}{0.15cm}
\includegraphics[width=0.9\textwidth]{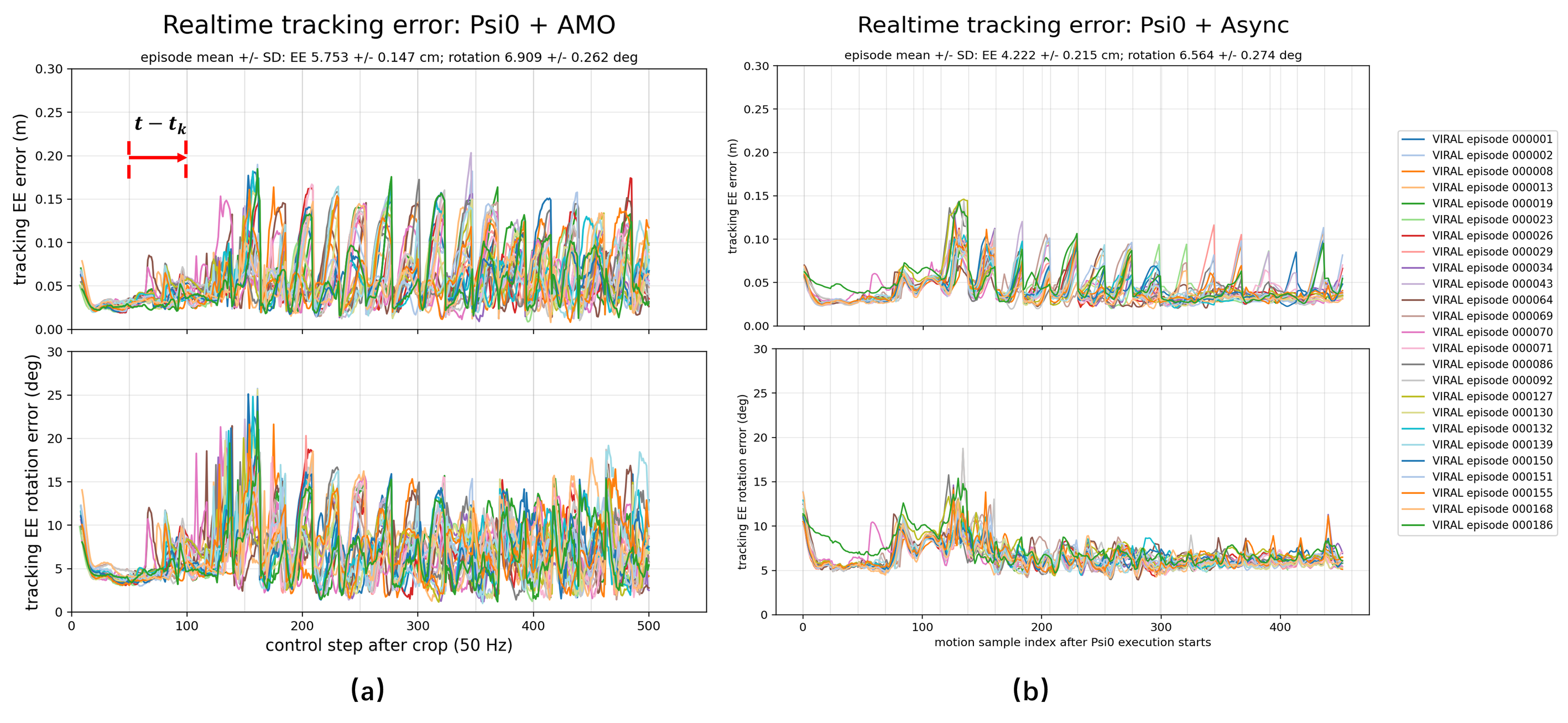}
\caption{Tracking errors under identical $\Psi_0$ planner outputs and a $0.46$-s update interval. Left: $\Psi_0$+AMO. Right: $\Psi_0$+ASYNC-CA. The upper and lower rows show end-effector position and rotation errors, respectively.}
\label{fig:vla_tracking_comparison}
\end{figure*}

Table~\ref{tab:online_planning_main} reports the online-planning comparison. GR00T+ASYNC-CA achieves 12/25 successes and lower tracking errors than $\Psi_0$+AMO, which achieves 6/25 successes. Because the planners use different action interfaces and horizons, this comparison evaluates the complete online pipelines rather than isolating the controller alone. Figure~\ref{fig:vla_tracking_comparison} separately compares the controllers using identical $\Psi_0$ planner outputs and a $0.46$-s update interval; ASYNC-CA exhibits lower and more stable tracking errors than AMO.

\section{Additional Hardware Demonstrations}\label{real_appendix}
For quantitative hardware evaluation, we attached retroreflective markers to the robot base and end-effectors and used a motion-capture system to measure the executed trajectories.

Table~\ref{tab:diverse_motion_statistics} reports simulation results on 14 motions and five hardware trials per motion (70 trials in total). ``Successful'' and ``All'' denote successful-motion and all-motion simulation statistics; hardware success denotes completion without falling. Smooth$_{\mathrm{real}}$ is the temporal mean of the joint-wise RMS acceleration from 17 measured upper-body joints.

\begin{table*}[!t]
\centering
\caption{Diverse 14-motion simulation evaluation and hardware deployment of all 14 motions.}
\label{tab:diverse_motion_statistics}
\resizebox{0.94\textwidth}{!}{%
\begin{tabular}{llccccc}
\toprule
Policy & Statistics & Success & \makecell{Pos$_{1\mathrm{s}}$\\(cm)} & \makecell{Rot$_{1\mathrm{s}}$\\($^\circ$)} & Joint margin & \makecell{Smooth$_{\mathrm{real}}$\\(rad/s$^2$)} \\
\midrule
Before PT & Successful & $5/14$ (35.70\%) & 10.37 & 11.85 & 0.13 & 8.06 \\
 & All & -- & 12.46 & 13.27 & 0.15 & 13.35 \\
\midrule
After PT & Successful & $12/14$ (85.70\%) & 11.11 & 12.00 & 0.14 & 10.72 \\
 & All & -- & 11.27 & 12.15 & 0.14 & 11.50 \\
 & Hardware & $60/70$ (85.70\%) & 13.42 & 14.32 & -- & 10.52 \\
\midrule
W/o guidance & Successful & $6/14$ (42.90\%) & 12.63 & 13.90 & 0.11 & 11.55 \\
 & All & -- & 12.73 & 13.59 & 0.12 & 14.47 \\
\midrule
SONIC w. ASYNC & Successful & $6/14$ (42.90\%) & 25.69 & 14.59 & 0.11 & 8.87 \\
 & All & -- & 28.10 & 18.13 & 0.13 & 11.81 \\
\bottomrule
\end{tabular}}
\end{table*}

Table~\ref{tab:matched_hardware} gives the absolute values underlying Fig.~\ref{fig:post-train-real}(a). Each policy executes the same five motions, selected by before-PT simulation success, five times each. Success is fall-based and tracking errors use motion capture. The radar score is $s=0.12+0.88(x_{\max}-x)/(x_{\max}-x_{\min})$ across the four policies; larger radii therefore mean lower errors or acceleration, not higher success.

\begin{table*}[!t]
\centering
\caption{Matched hardware comparison: five motions, five trials each.}
\label{tab:matched_hardware}
\begin{tabular}{lcccc}
\toprule
Policy & Succ. & \makecell{Pos$_{1\mathrm{s}}$\\(cm)} & \makecell{Rot$_{1\mathrm{s}}$\\($^\circ$)} & \makecell{Smooth$_{\mathrm{real}}$\\(rad/s$^2$)} \\
\midrule
Before PT & $25/25$ & 16.28 & 16.31 & 9.09 \\
After PT & $25/25$ & 13.24 & 13.22 & 9.48 \\
W/o guidance & $21/25$ & 19.03 & 24.10 & 14.01 \\
SONIC w. ASYNC & $15/25$ & 19.64 & 20.71 & 9.30 \\
\bottomrule
\end{tabular}
\end{table*}

\begin{figure*}[!t]
    \centering
    \includegraphics[width=0.9\textwidth,height=0.78\textheight,keepaspectratio]{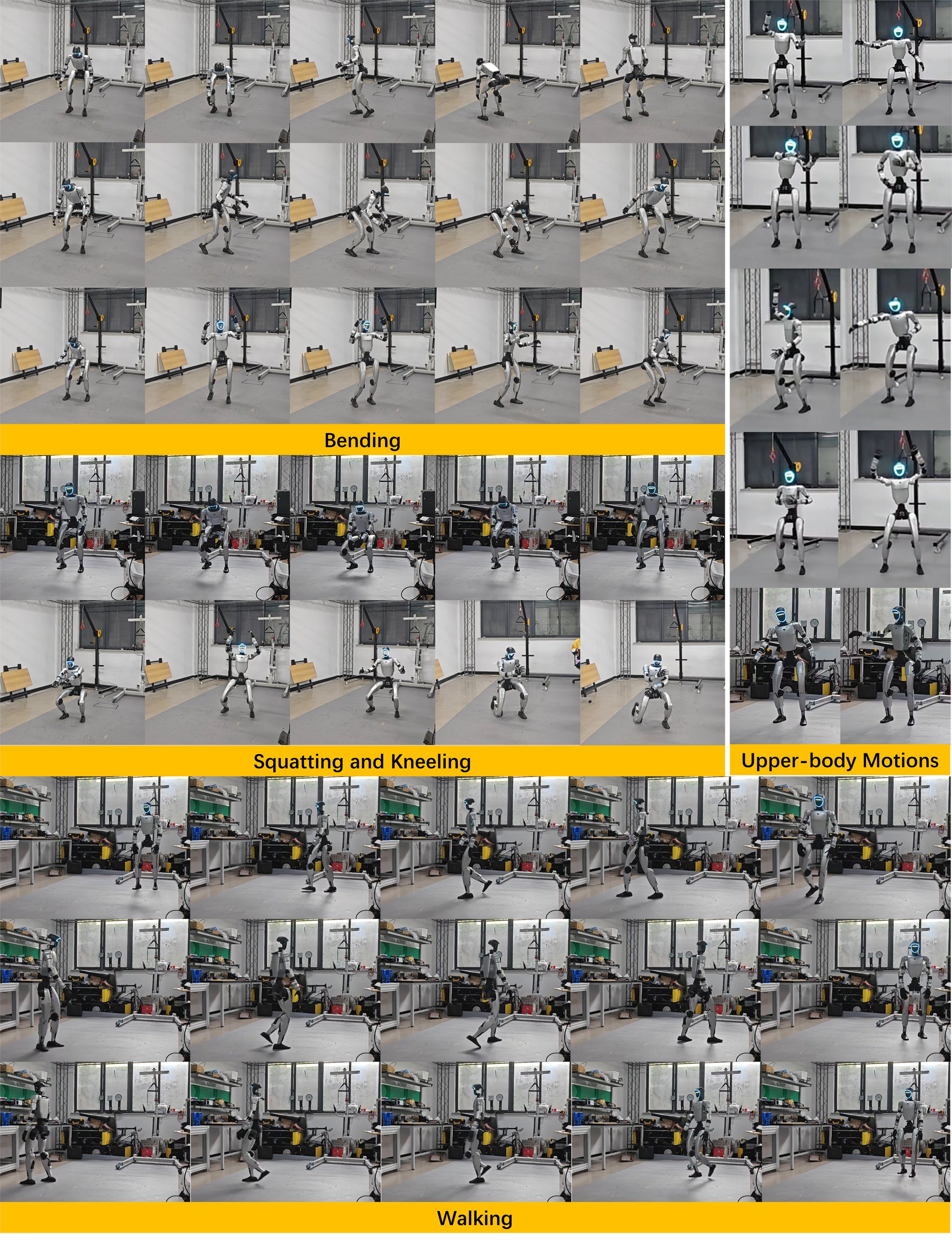}
    \caption{Additional hardware demonstrations on the Unitree G1. The post-trained policy executes diverse whole-body motions including waving hands, object manipulation, squatting, and walking.}
    \label{fig:additional-hardware}
\end{figure*}

Figure~\ref{fig:additional-hardware} further shows upper-body gestures, object manipulation, deep squatting, kneeling, and walking driven by sparse upper-body trajectory references.

\end{document}